\documentclass[10pt,twocolumn,letterpaper]{article}
\usepackage[pagenumbers]{cvpr}
\usepackage[T1]{fontenc}
\usepackage[utf8]{inputenc}
\usepackage[english]{babel}
\usepackage{wrapfig}
\usepackage{multirow}
\usepackage{makecell}
\usepackage{cuted}
\usepackage{float}
\usepackage[accsupp]{axessibility}
\usepackage[autostyle]{csquotes}
\usepackage{microtype}
\DeclareUnicodeCharacter{2011}{\mbox{-}}
\definecolor{cvprblue}{rgb}{0.21,0.49,0.74}
\usepackage[breaklinks,colorlinks,allcolors=cvprblue]{hyperref}
\usepackage{orcidlink}
\def\paperID{12751}
\def\confName{CVPR}
\def\confYear{2026}

\makeatletter
\patchcmd{\@maketitle}
    {\vspace*{12pt}}
    {\vspace*{0pt}}
    {}
    {}
\makeatother
\begin{document}

\title{Novel View Synthesis using DDIM Inversion}
\author{%
Sehajdeep Singh\thanks{%
Work performed while affiliated with Indraprastha Institute
of Information Technology, Delhi.
}
\qquad
A~V~Subramanyam
\qquad
Aditya Gupta
\qquad
Sahil Gupta
\\
Indraprastha Institute of Information Technology, Delhi
\\
{\small \texttt{sehajsasan@gmail.com}\textsuperscript{*}}
\\
{\small\ttfamily
\{subramanyam,aditya22031,sahil22430\}@iiitd.ac.in}
}

\maketitle

\begin{strip}
    \centering
    \newcommand{\teaserheight}{8.5cm}

    \begin{minipage}[t]{0.68\textwidth}
        \centering
        \includegraphics[
            width=\linewidth,
            height=\teaserheight
        ]{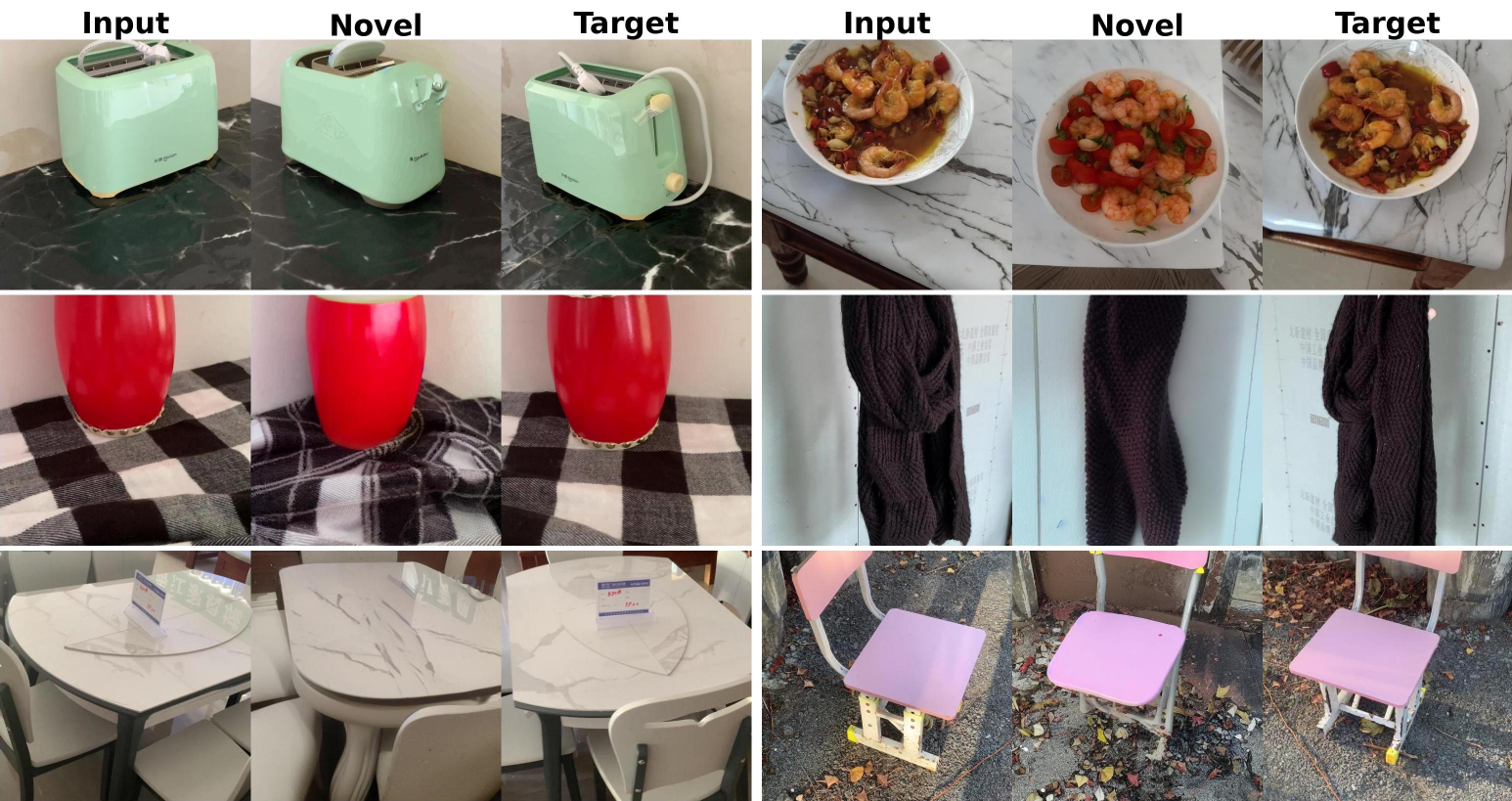}
        \small (a) MvImgNet test set
    \end{minipage}%
    \hfill
    \begin{minipage}[t]{0.30\textwidth}
        \centering
        \includegraphics[
            width=\linewidth,
            height=\teaserheight
        ]{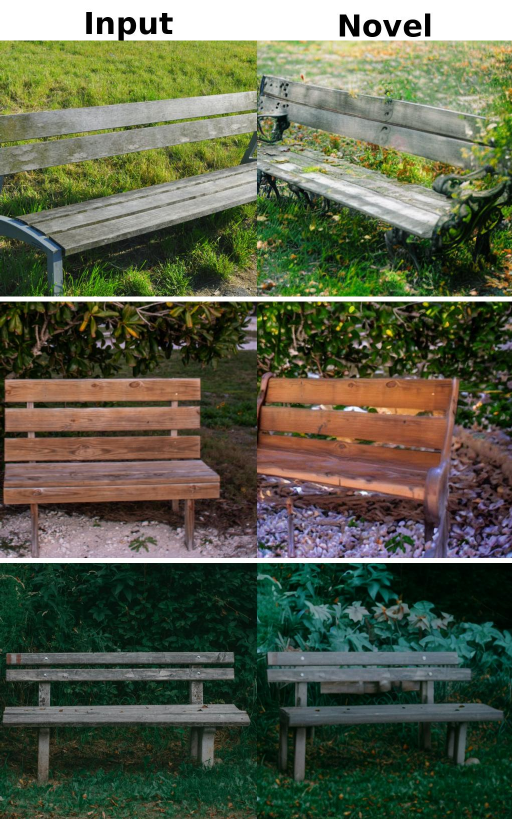}
        \small (b) OOD images
    \end{minipage}

    \captionof{figure}{%
        (a) High-resolution ($512\times512$) novel-view synthesis
        on the MvImgNet test set from a single input image and
        camera parameters (b) Zero-shot synthesis on out-of-domain
        images downloaded from Unsplash.%
    }
    \label{fig:teaser}
\end{strip}

\begin{abstract}
Single-image novel view synthesis requires viewpoint-dependent structural
changes while retaining source-scene information and inferring unobserved
content. We formulate this task as a camera-conditioned   initialization problem for a
frozen pretrained latent diffusion model. At an intermediate DDIM inversion
timestep, we decompose the source representation into a structural mean and
a source noise residual. A pose-conditioned translation U-Net
(TUNet) learns to predict the target-view mean from the source mean. We then
introduce a score-guided fusion strategy: the predicted mean and source
noise residual form a proposal that is evaluated by the frozen denoiser, whose
noise prediction is combined with the translated mean to initialize DDIM
sampling. Only
TUNet is trained, while the pretrained model synthesizes the final view
without explicit reconstruction or rendering of a shared 3D scene.
Experiments on MVImgNet and RealEstate10K demonstrate competitive
performance.
Beyond image-level metrics, Wasserstein comparisons show that the corrected
initialization yields trajectory-curvature distributions closer to those
of ground-truth target-initialized sampling than raw fusion. These results
support structural translation and source-conditioned initialization as
a practical approach to reusing frozen diffusion priors for novel view
synthesis.
The code is available \href{https://github.com/Visual-Conception-Group/ddim_nvs}
{here}. 
\end{abstract} 

\section{Introduction}

Novel view synthesis is a fundamental task in computer vision and graphics. 
Remarkable works such as NeRFs \cite{mildenhall2020nerfrepresentingscenesneural} and 3DGS \cite{kerbl20233dgaussiansplattingrealtime} are extensively used in 3D scene understanding. 
Several works improve upon these foundational methods. However, their dependence on scene-level optimization and the need for a dense set of views limit usability. 
Diffusion models \cite{Rombach_2022_CVPR,podell2023sdxlimprovinglatentdiffusion} have  gained significant traction for novel view synthesis \cite{elata2025novel,tang2024cycle3dhighqualityconsistentimageto3d,liang2025wonderland,zhou2025stable,zhang2025ar}. 
A classical approach is to fine-tune these models on 3D datasets along with a module that encodes geometric priors into the architecture \cite{tang2023mvdiffusionenablingholisticmultiview,liu2024syncdreamergeneratingmultiviewconsistentimages,long2023wonder3dsingleimage3d,huang2024epidiffenhancingmultiviewsynthesis,bourigault2024mvdiffscalableflexiblemultiview,gao2024cat3dcreate3dmultiview}. 
However, despite producing plausible images, such approaches often suffer from structural inconsistencies across views, especially for long-range viewpoint transformations.

DDIM \cite{song2022denoisingdiffusionimplicitmodels} proposed a deterministic inversion procedure, \enquote{DDIM Inversion}, which sequentially adds noise to a image latent to obtain a close to gaussian noise representation of the latent. The said noisy latent can be retraced to the original image using DDIM sampling. This latent encapsulates both the signal and the noise components of the noisy representation whose distribution evolves over inversion time. 
 Recent works \cite{garibi2024renoiserealimageinversion,mokady2022nulltextinversioneditingreal,staniszewski2024there} observe that the DDIM inversion space is less manipulable and difficult to interpolate directly for downstream tasks such as editing or view transformation. 

In this work, we take a different perspective to view synthesis. Rather than treating novel view synthesis as an image prediction or diffusion conditioning problem, we reformulate it as a diffusion trajectory initialization problem within a frozen, pretrained diffusion model. Our key observation is that DDIM inversion at intermediate timesteps produces latents with a meaningful internal structure: a deterministic mean component that captures low-frequency scene geometry, and a stochastic variance component that governs high-frequency appearance. This decomposition enables us to decouple pose-induced structural transformation from appearance synthesis, allowing view translation to be performed directly in the DDIM-inverted mean space while relying on the pretrained diffusion dynamics to restore fine-grained details.

We first learn to map the input-view inverted mean latent at a fixed intermediate diffusion timestep to its corresponding target-view latent using a lightweight translation network (TUNet). This latent primarily encodes the deterministic signal component of the scene at that noise level. The learned mapping produces a coarse structural approximation of the target view, consistent with the known spectral bias of diffusion models toward low-frequency components \cite{choi2022perception}. 
However, directly using this translated latent for sampling often leads to unstable diffusion evolution and insufficient high-frequency reconstruction. 
To address this, we propose a variance-guided noisy latent fusion strategy that re-aligns the initialization of the diffusion trajectory. We utilise the presence of structured correlations in DDIM inversion noise to formulate the proposed variance guided fusion strategy. Beyond injecting high-frequency details, the proposed fusion stabilizes the latent initialization so that the resulting sampling trajectory exhibits smoother and more consistent evolution during diffusion sampling. In our experiments, we show that the fused latent leads to diffusion paths that more closely resemble those obtained from true target-view sampling trajectories, resulting in improved structural consistency and image quality.

Notably, we use a pretrained diffusion model and only train a lightweight latent-space translation network (TUNet). Sample results are shown in Figure~\ref{fig:teaser}. We claim the following contributions:





\begin{itemize}
\item Pose-Conditioned Structural Translation in Diffusion Mean Space.
We propose TUNet, a lightweight pose-conditioned translation network that performs view transformation on the structural mean component of intermediate DDIM latents, effectively capturing viewpoint-induced geometric changes while remaining agnostic to high-frequency appearance details.

\item Score Guided and Diffusion-Dynamics aligned Latent Fusion.
We present a principled fusion strategy that exploits structured noise correlations in DDIM inversion and uses the pretrained diffusion score network to propose the target view DDIM Inversion initialization latent that constructs the target view stabilizing sampling and enabling high-fidelity detail reconstruction.

\item Trajectory-Level Analysis of Diffusion View Synthesis.
We provide an in-depth analysis of diffusion sampling trajectories in \ref{sec:curvature_analysis}, showing that our fusion strategy produces second-order trajectory statistics that more closely match ground-truth target-view trajectories, offering a new perspective on evaluating diffusion-based view synthesis beyond perceptual and reconstruction image metrics.

\end{itemize}

\section{Related Works}

\noindent{\textbf{Transformers}:} 
GeoGPT \cite{rombach2021geometry} was one of the early works to perform view synthesis using transformers. 
NViST \cite{jang2024nvistwildnewview} adopts a transformer-based encoder-decoder architecture \cite{vaswani2023attentionneed,dosovitskiy2021imageworth16x16words} to predict a radiance field from a single image, enabling novel view synthesis via NeRF-style volumetric rendering. 

However, NViST suffers from loss of fine details due to aggressive downsampling (by a factor of 12), and it struggles to synthesize long-range viewpoints (when the target frame is more than 15 frames away from the input frame in a 30‑frame sweep).

\noindent{\textbf{Diffusion models}}: Diffusion models can be leveraged to generate plausible content in the unobserved regions of the input views. In the following, we identify them as the ones which finetune pre-trained diffusion models, or train diffusion models from scratch.

\noindent{\textbf{Pretrained Diffusion Models}}: MVDiffusion \cite{tang2023mvdiffusionenablingholisticmultiview}, Zero123++ \cite{shi2023zero123singleimageconsistent},  SyncDreamer \cite{liu2024syncdreamergeneratingmultiviewconsistentimages}, Wonder3D \cite{long2023wonder3dsingleimage3d}, EpiDiff \cite{huang2024epidiffenhancingmultiviewsynthesis}, BoostDream \cite{Yu_2024}, MVDiff \cite{bourigault2024mvdiffscalableflexiblemultiview}, CAT3D \cite{gao2024cat3dcreate3dmultiview}, Magic-Boost \cite{yang2025magicboostboost3dgeneration}, Cycle3D \cite{tang2024cycle3dhighqualityconsistentimageto3d}, GenWarp \cite{seo2024genwarp}, use a pretrained or finetuned diffusion model. MVDiffusion modifies the Stable Diffusion architecture by introducing a cross-branch attention mechanism, known as correspondence-aware attention (CAA), to model inter-view dependencies. SyncDreamer constructs a view frustum feature volume from all the target noisy views and injects these into the pretrained denoising Unet using depth-wise attention layers. EpiDiff \cite{huang2024epidiffenhancingmultiviewsynthesis} integrates an attention module guided by epipolar constraints into the intermediate and decoding stages of the U-Net, enabling the model to capture generalized epipolar geometry across views. GenWarp \cite{seo2024genwarp} introduces a warp and inpaint technique.
Most of these existing approaches either fine-tune the diffusion model or inject spatial features corresponding to the target view into the base model’s denoising U-Net. Such features are typically derived from volumetric projections or depth estimates. However, models that follow this paradigm often struggle with scene-level reconstructions and are usually trained on object-specific datasets, which may limit the generalization to complex scenes.

In contrast, our method does not modify or inject any learned features into the U-Net of the diffusion model. Instead, we provide external conditioning input to TUNet to obtain the latent corresponding to the target view.

\begin{figure*}[t]
    \centering
 
    \includegraphics[width=1\textwidth]{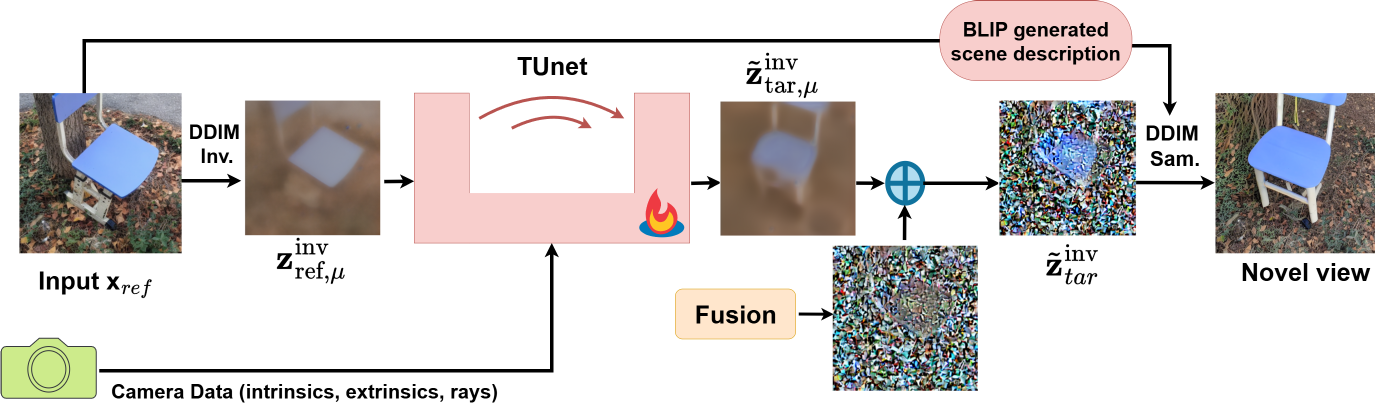}
    \caption{ Overview: Given a single reference image $\mathbf{x_{\text{ref}}}$, we first apply DDIM inversion up to \(t=600\) to obtain the mean latent \(\mathbf{z}_{\text{ref},\mu}^{\text{inv}}\). This, together with camera intrinsics/extrinsics, class embeddings, and ray information, is fed into our translation network \textbf{TUNet}. TUNet predicts the target-view mean latent \(\tilde{\mathbf{z}}_{\text{tar},\mu}^{\text{inv}}\), which we combine with the corresponding noise component using our fusion strategy to form the initial DDIM latent \(\tilde{\mathbf{z}}_{tar}^{\text{inv}}\). Finally, this latent is used for sampling by a pre-trained diffusion model to synthesize the novel view image.}
    \label{fig:pipeline}
\end{figure*}

\noindent \textbf{Training Diffusion Model from Scratch}: Several recent works train diffusion models from scratch for novel view synthesis, including \cite{tseng2023consistentviewsynthesisposeguided}, Photometric-NVS \cite{yu2023longtermphotometricconsistentnovel}, DiffDreamer \cite{cai2023diffdreamerconsistentunsupervisedsingleview}, GIBR \cite{anciukevivcius2024denoising}, \cite{henderson2024sampling3dgaussianscenes}, and VIVID \cite{elata2025novel}.
Photometric-NVS \cite{yu2023longtermphotometricconsistentnovel} introduces a two-stream latent diffusion architecture that independently processes the source and noisy target views, while exchanging information via pose-conditioned cross-attention mechanisms. GIBR \cite{anciukevivcius2024denoising} models 3D scenes using IB-planes and trains the diffusion process directly in pixel space, enabling learning of a joint distribution over multi-view observations and camera poses. VIVID \cite{elata2025novel} trains diffusion model in pixel-space.

Training entire diffusion models end-to-end is computationally expensive and requires large-scale datasets to achieve high-resolution and photorealistic reconstruction.
In contrast, our method operates in the DDIM-inverted latent space at a fixed timestep, which corresponds to a weak yet informative signal. This allows us to perform translation from a given latent to a target latent using a lightweight translation U-Net. Operating in the DDIM Inversion latent space significantly simplifies the view translation task, as the model works with compact, semantically rich representations rather than raw pixels.

\section{Method}
\label{sec:method}

We address single-image novel view synthesis by operating entirely in the DDIM-inverted latent space of a pretrained latent diffusion model (LDM)~\cite{Rombach_2022_CVPR}. 
Our method consists of two main stages. First, we learn a lightweight translation network (TUNet) that maps the input-view reference inverted latent's mean at a fixed intermediate timestep to the corresponding target-view latent mean at the same timestep. Second, the translated latent mean is fused with a score corrected noise residual and this fused latent which combines target view mean prediction and the score corrected noise residual is used to initialize DDIM sampling with a pretrained score network to reconstruct the final novel view.

Our motivation arises from the fact that at an intermediate DDIM timestep, the inverted latent contains a dominant structural component along with a stochastic component. We perform viewpoint translation on the structural component while leveraging the pretrained diffusion model during sampling to restore high-frequency details and maintain consistency with the learned diffusion prior. An overview of the pipeline is shown in Figure~\ref{fig:pipeline}.

    \subsection{Spectral Behavior of Diffusion and DDIM Mean-Variance Decomposition}
\label{sec:diffusion_prelims}
\begin{figure}[t]
    \centering

    \begin{minipage}{0.52\linewidth}
        \centering
        \includegraphics[width=0.48\linewidth]{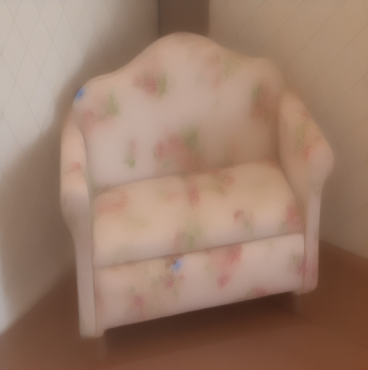}\hfill
        \includegraphics[width=0.48\linewidth]{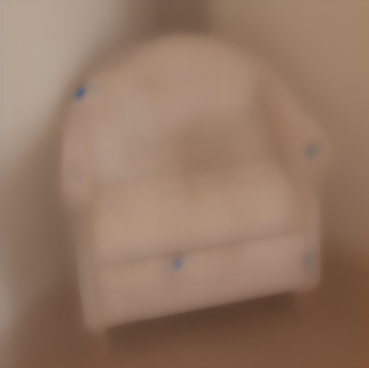}\\[2pt]
        \includegraphics[width=0.48\linewidth]{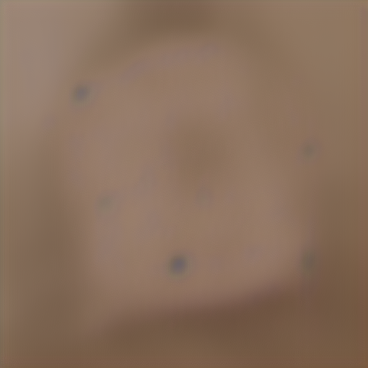}\hfill
        \includegraphics[width=0.48\linewidth]{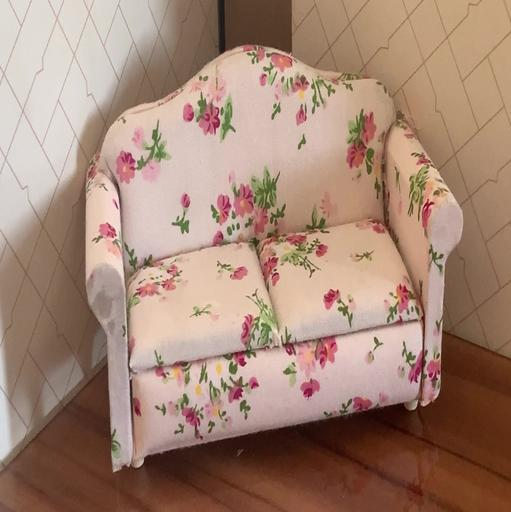}
    \end{minipage}

    \caption{Mean of the DDIM inverted latent.
    Top: $t=400, 600$. Bottom: $t=800$ and original image.}
    \label{fig:dataset_example}
\end{figure}
The forward diffusion process~\cite{ho2020denoising} is given by
$
\mathbf{x}_t =
\sqrt{\overline{\alpha}_t}\,\mathbf{x}_0
+
\sqrt{1-\overline{\alpha}_t}\,\boldsymbol{\epsilon},
\quad
\boldsymbol{\epsilon} \sim \mathcal{N}(\mathbf{0}, \mathbf{I})$,
where $\mathbf{x}_0$ is the clean latent and $\mathbf{x}_t$ is the latent at timestep $t$. 
Recent analyses~\cite{kingma2024understanding,falck2025fourierspaceperspectivediffusion,choi2022perception} show that diffusion exhibits a {spectral bias}, that is, high-frequency components are corrupted earlier in the forward process and recovered later in the reverse process. Consequently, at intermediate timesteps, the latent retains dominant low-frequency structural information, while high-frequency details are largely encoded in the noise term.
Using DDIM inversion~\cite{song2022denoisingdiffusionimplicitmodels}, the latent at timestep $t+1$ can be written as
\begin{equation*}
\mathbf{z}^{\text{inv}}_{t+1}
=
\underbrace{\mathbf{z}^{\text{inv}}_{\mu,t+1}}_{\text{structural mean}}
+
\underbrace{\mathbf{z}^{\text{inv}}_{\sigma,t+1}}_{\text{variance / noise}},
\end{equation*}
where $\mathbf{z}^{\text{inv}}_{\mu,t+1}$ corresponds to the signal (mean) term and $\mathbf{z}^{\text{inv}}_{\sigma,t+1}$ corresponds to the variance (noise) term. We interpret $\mathbf{z}^{\text{inv}}_{\mu}$ as a {coarse structural anchor}, and
$\mathbf{z}^{\text{inv}}_{\sigma}$ as encoding high-frequency appearance statistics.
Rather than inverting all the way to $t=T$, where the latent approaches white noise and reverse trajectories become unstable~\cite{bao2025freeinv,zeng2025inverting,feng2024wave}, we stop inversion at an intermediate timestep $t^* = 600$. Figure~\ref{fig:dataset_example} shows decoded mean latents at different timesteps. We observe that at $t=400$, the mean reflects dominant low frequencies which precludes the generation of diverse images. At $t=800$, the low frequency component is extremely weak. At $t=600$, the mean provides a coarse yet structurally coherent representation suitable for viewpoint translation. Since viewpoint transformation predominantly involves reorganization of spatial structure and scene geometry while fine-scale appearance details and disoccluded regions can be resolved during diffusion sampling, we perform pose-conditioned translation in the mean space and rely on the stochastic component to recover high-frequency details.



\subsection{Pose-Conditioned Structural Translation via TUNet}
\label{sec:tunet_transport}

Let $\mathbf{z}^{\mathrm{inv}}_{\mathrm{ref},\mu}$ denote the DDIM-inverted mean latent of the input view reference image at timestep $t^* = 600$, and let 
$C_{\mathrm{ref}}$ and $C_{\mathrm{tar}}$ denote the reference and target camera parameters.

\noindent \textbf{Structural Translation in Mean Space.}
We learn a pose-conditioned mapping
\begin{equation}
TUNet :
\big(
\mathbf{z}^{\mathrm{inv}}_{\mathrm{ref},\mu},
C_{\mathrm{ref}},
C_{\mathrm{tar}}
\big)
\longrightarrow
\tilde{\mathbf{z}}^{\mathrm{inv}}_{\mathrm{tar},\mu},
\label{eq:tunet_mapping}
\end{equation}
which predicts the DDIM-inverted mean latent corresponding to the target view conditioned on camera parameters. Thus, TUNet approximates a structural translation operator acting on the low-frequency diffusion manifold.

\noindent \textbf{Architecture and Conditioning.}
TUNet follows a U-Net encoder–decoder architecture. 
The input is the reference mean latent $\mathbf{z}^{\mathrm{inv}}_{\mathrm{ref},\mu}$. Camera parameters are embedded by vectorizing intrinsics and extrinsics, $C = (K, R, t)$,
and projecting them through a learnable linear layer to obtain an embedding vector $\mathbf{e}_C$. 
A learnable class embedding $\mathbf{e}_c$ is also used to preserve semantic consistency. 
These embeddings are concatenated with the diffusion timestep embedding $\boldsymbol{\gamma}(t^*)$ and projected to match the feature dimensionality:
\[
\mathbf{e}_{\mathrm{cond}}
=
\mathrm{Proj}
\big[
\boldsymbol{\gamma}(t^*)
\oplus
\mathbf{e}_C
\oplus
\mathbf{e}_c
\big].
\]

Conditioning is injected additively at multiple stages of the network. 
The encoder blocks are conditioned on the reference camera parameters, preserving structural encoding of the input view. 
The bottleneck is jointly conditioned on both reference and target cameras, enabling viewpoint transition at the latent level. 
The decoder blocks are conditioned only on the target camera parameters, guiding the representation toward the desired pose.

\noindent \textbf{Cross-View Feature Alignment.}
To enforce geometric consistency during translation, we integrate cross-attention between reference and target representations in the bottleneck and upsampling stages.
Ray embeddings are computed from camera parameters using standard ray parameterization (as in NeRF \cite{yu2021pixelnerfneuralradiancefields}). 
Let $\mathbf{r}_{\mathrm{ref}}$ and $\mathbf{r}_{\mathrm{tar}}$ denote the ray embeddings of the reference and target views, respectively. 
Given intermediate target features $\mathbf{f}_{\mathrm{tar}}$, we define:
\[
\begin{gathered}
\mathbf{Q} = \mathbf{W}_Q [\mathbf{r}_{\mathrm{tar}} \mathbin{\|} \mathbf{f}_{\mathrm{tar}}],
\\
\mathbf{K} = \mathbf{W}_K [\mathbf{r}_{\mathrm{ref}} \mathbin{\|} \mathbf{z}^{\mathrm{inv}}_{\mathrm{ref},\mu}],
\\
\mathbf{V} = \mathbf{W}_V \mathbf{z}^{\mathrm{inv}}_{\mathrm{ref},\mu}.
\end{gathered}
\]

Attention is computed as, $
\mathrm{Attn}(\mathbf{Q},\mathbf{K},\mathbf{V})
=
\mathrm{softmax}
\left(
\frac{\mathbf{Q}\mathbf{K}^\top}{\sqrt{d}}
\right)
\mathbf{V}$,
and added to the target features. 

This mechanism enables pose-aware feature transfer from the reference mean latent to the target mean representation. 
Importantly, cross-attention operates directly in mean latent space, reinforcing the fact that translation acts on structural components rather than texture-level noise.


\noindent \textbf{Training Objective.}
We supervise TUNet using an $\ell_2$ loss between the predicted and ground-truth target mean latents:
\begin{equation}
\mathcal{L}_{\mathrm{MSE}}
=
\left\|
\tilde{\mathbf{z}}^{\mathrm{inv}}_{\mathrm{tar},\mu}
-
\mathbf{z}^{\mathrm{inv}}_{\mathrm{tar},\mu}
\right\|_2^2,
\label{eq:mse_loss}
\end{equation}
where $\mathbf{{z}}^{\mathrm{inv}}_{\mathrm{tar},\mu}$ is obtained by extracting the DDIM inverted mean term of the ground-truth target latent at timestep $t^*$.
\subsection{Diffusion Dynamics Aligned Fusion   }
\label{sec:fusion_projection}

After pose-conditioned structural translation, we obtain the predicted target mean latent 
$\tilde{\mathbf{z}}^{\text{inv}}_{\text{tar},\mu}$ at timestep $t^*$. 
To reintroduce appearance statistics from the reference view, we construct a fused latent:
\begin{equation}
\mathbf{z}_{\text{fused}}
=
\tilde{\mathbf{z}}^{\text{inv}}_{\text{tar},\mu}
+
\mathbf{z}^{\text{inv}}_{\text{ref},\sigma}.
\label{eq:fused_latent}
\end{equation}

Here, $\mathbf{z}^{\text{inv}}_{\text{ref},\sigma}$ denotes the variance (noise) component of the DDIM-inverted input view reference latent at the same timestep $t^*$. 
This fusion produces a latent that combines (i) target view pose-aligned structural information from TUNet, and
(ii) reference or input view specific high-frequency statistics.

We do not assume that  $\mathbf{z}^{\text{inv}}_{\text{ref},\sigma}$ in Eq. \ref{eq:fused_latent} is geometrically
aligned with the target view. For large viewpoint changes, this is generally not
guaranteed. Instead, $\mathbf{z}_{\mathrm{fused}}$ is treated as a
\emph{source-conditioned stochastic proposal}: it combines target-view structure
with source-scene residual statistics.

\textbf{Correcting the proposed $\mathbf{z}_{\text{fused}}$ with the frozen score model}: We evaluate the proposed $\mathbf{z}_{\text{fused}}$ using the pretrained diffusion U-Net at the same
timestep:
\begin{equation}
    \epsilon_\phi
    =
    \mathrm{UNet}_\phi(z_{\mathrm{fused}}, t^*).
    \label{eq:noise_prediction}
\end{equation}
The predicted noise $\epsilon_\phi$ is a score-related correction under the
pretrained diffusion model. It estimates how the proposed latent $\mathbf{z}_{\text{fused}}$ should be
adjusted to better align with the learned noised data distribution at timestep
$t^*$ from the target viewpoint. We then construct the final target-view sampling initialization:
\begin{equation}
    \tilde{z}^{\mathrm{inv}}_{\mathrm{tar}}
    =
    \tilde{z}^{\mathrm{inv}}_{\mathrm{tar},\mu}
    +
    \bar{\sigma}_{t^*}\epsilon_\phi,
    \label{eq:final_latent}
\end{equation}
where $\bar{\sigma}_{t^*}=\sqrt{1-\bar{\alpha}_{t^*}}$ is the marginal noise scale at timestep $t^*$.
In score-based diffusion models, the noise predictor $\boldsymbol{\epsilon}_\phi(\mathbf{z}, t)$ is proportional to the score $\nabla_{\mathbf{z}} \log p_t(\mathbf{z})$,
that is, the gradient of the log-density of the noised data distribution at timestep $t$.

Importantly, the predicted noise $\boldsymbol{\epsilon}_\phi$ is {not} a direct reuse of the reference variance term. 
Instead, it is a pose-adaptive score correction computed from the fused latent under the pretrained score network. 

Hence, in our pipeline the fused latent $\mathbf{z}_{\text{fused}}$ is first evaluated by the frozen diffusion U-Net, and the resulting noise prediction replaces the initial variance term before DDIM sampling. 
This ensures that the final target-view sampling initialization $\tilde{\mathbf{z}}^{\text{inv}}_{\text{tar}}$ lies closer to the high-density region of the learned diffusion manifold at timestep $t^*$ biased towards the target viewpoint.


\subsection{Inference}
The latent $\tilde{\mathbf{z}}^{\text{inv}}_{\text{tar}}$ is used as the starting point for DDIM sampling to generate the final novel view.
Our method does not explicitly model 3D geometry, or multi-view consistency constraints. Instead, geometric plausibility emerges implicitly from pose-conditioned structural translation in DDIM mean space,
 as well as regularization via the pretrained diffusion manifold. 
 

\begin{figure*}[!t]
    \centering
    \includegraphics[width=.95\textwidth]{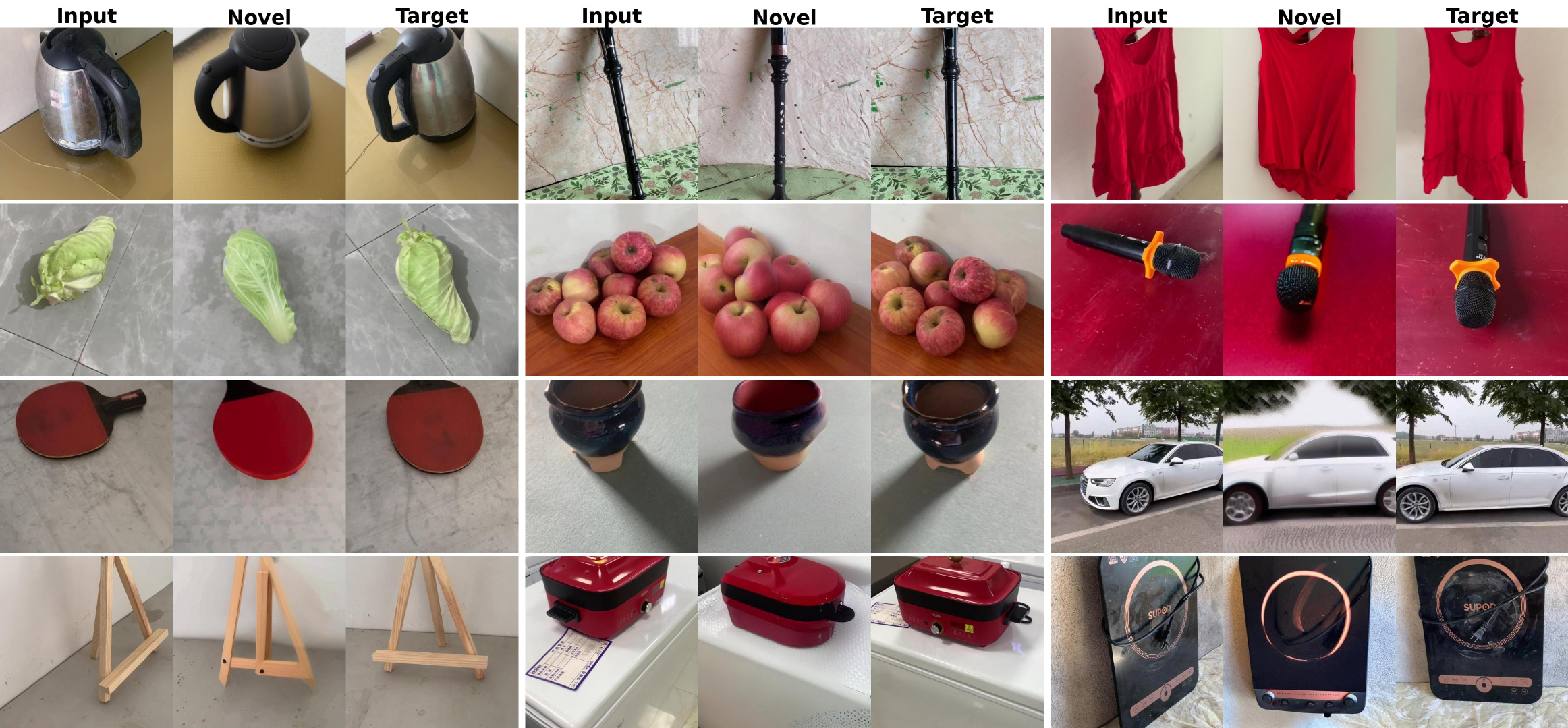}
    \caption{Qualitative results with our 167-class trained model. }
    \label{fig:grid_2}
\end{figure*}

\begin{figure*}[t]
    \centering
    \includegraphics[height=0.88\textheight,width=\textwidth,keepaspectratio]{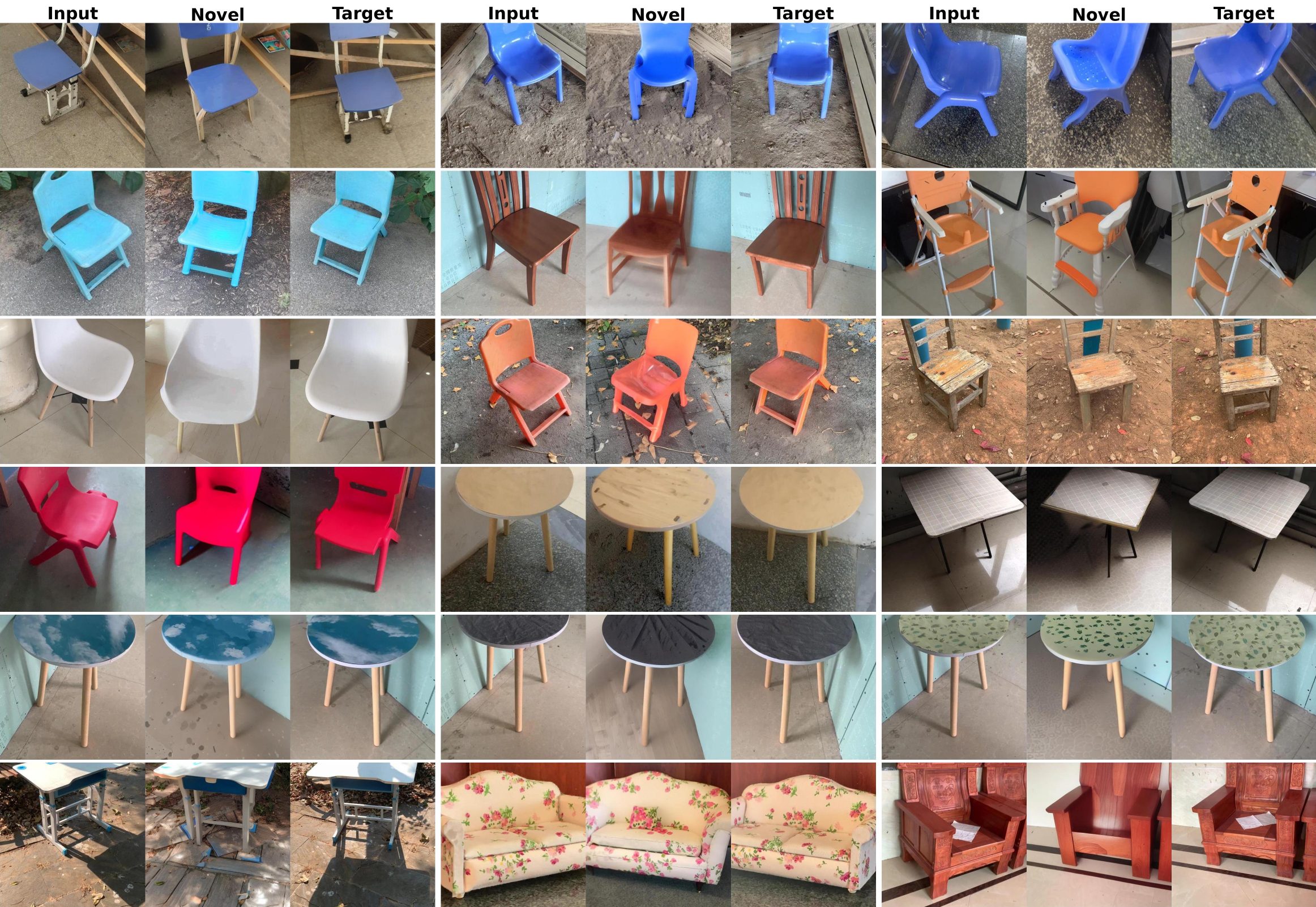}
    \caption{Qualitative results with our 3-class trained model on MVImgNet test set. }
    \label{fig:grid_3_cls}
\end{figure*}

\section{Experiments}
\label{sec:Experiments}

\textbf{Dataset}: We perform experiments using MvImgNet \cite{yu2023mvimgnetlargescaledatasetmultiview} and RealEstate10K \cite{zhou2018stereo}. MvImgNet consists of 6.5
million frames of real-world scenes across 238 categories. We use two subsets of MVImgNet (i) three scene categories: sofas, chairs, and tables. A 90-5-5 \cite{anciukevivcius2024denoising} split is used for training, validation, and testing, respectively, determined by lexicographic ordering of the scene identifiers.
(ii) We use 8.5 lakh frames across 167 classes and for each class, we keep 1 scene out of 99 in the test set to evaluate our results and compare with other methods. In case of RealEstate10K, we train using 1 million pairs.

We compare our 3-class model with GIBR \cite{anciukevivcius2024denoising} and the 167-class model with NViST \cite{jang2024nvistwildnewview}. For GIBR, we have the same train/test split and directly report the results from their paper. For NViST, we use their pre-trained model to test exact input/target frame pairs. All of the testing frames are from unseen scenes within the classes used in training. To compare with GIBR, we resize our $512\times512$ results to $256\times256$. For direct comparison with NViST, we resize our results to $90\times90$. We report LPIPS and FID scores with the resized results for comparisons. We follow the evaluation protocol as given in \cite{anciukevivcius2024denoising,denton2018stochastic}.


\subsection{Trajectory Curvature Analysis}
\label{sec:curvature_analysis}

\begin{table*}[t]
\centering
\caption{(a) Full-trajectory curvature Wasserstein distance to the ground-truth target diffusion trajectory (lower is better). (b) Bin-wise curvature analysis. FD is frame distance}
\vspace{0pt}
\begin{subtable}[t]{0.49\textwidth}
\centering
\caption{}
\vspace{0pt}
\label{tab:curvature_full}
\resizebox{0.84\linewidth}{!}{
\begin{tabular}{l@{\hspace{6pt}}c@{\hspace{6pt}}c}
\toprule
 & \multicolumn{2}{c}{Initial sampling latent} \\
\cmidrule(lr){2-3}
FD & $\tilde{\mathbf{z}}^{\mathrm{inv}}_{\mathrm{tar}}$ & $\mathbf{z}_{\text{fused}}$ \\
\midrule
0--10  & $0.0856 \pm 0.0467$ & $0.0904 \pm 0.0500$ \\
10--20 & $0.0662 \pm 0.0286$ & $0.0985 \pm 0.0368$ \\
20--25 & $0.0675 \pm 0.0372$ & $0.0848 \pm 0.0373$ \\
\bottomrule
\end{tabular}}
\end{subtable}\hspace{10pt}
\begin{subtable}[t]{0.45\textwidth}
\centering
\caption{}
\vspace{0pt}
\label{tab:curvature_temporal}
\resizebox{\linewidth}{!}{
\begin{tabular}{lcccccc}
\toprule
FD
& \shortstack{Initial\\ latent}
& \shortstack{Bin1\\601--501}
& \shortstack{Bin2\\481--381}
& \shortstack{Bin3\\361--261}
& \shortstack{Bin4\\241--141}
& \shortstack{Bin5\\121--0} \\
\midrule
\multirow{2}{*}{0--10}
& $\tilde{\mathbf{z}}^{\mathrm{inv}}_{\mathrm{tar}}$
& 0.0797 & 0.0550 & 0.0360 & 0.0281 & 0.0329 \\
& $\mathbf{z}_{\text{fused}}$
& 0.1065 & 0.0744 & 0.0552 & 0.0389 & 0.0335 \\
\midrule
\multirow{2}{*}{10--20}
& $\tilde{\mathbf{z}}^{\mathrm{inv}}_{\mathrm{tar}}$
& \textbf{0.0692} & \textbf{0.0180} & \textbf{0.0153} & \textbf{0.0167} & 0.0296 \\
& $\mathbf{z}_{\text{fused}}$
& 0.1144 & 0.0803 & 0.0441 & 0.0392 & 0.0292 \\
\midrule
\multirow{2}{*}{20--25}
& $\tilde{\mathbf{z}}^{\mathrm{inv}}_{\mathrm{tar}}$
& \textbf{0.0414} & \textbf{0.0300} & \textbf{0.0222} & \textbf{0.0161} & 0.0292 \\
& $\mathbf{z}_{\text{fused}}$
& 0.1118 & 0.0635 & 0.0340 & 0.0297 & 0.0358 \\
\bottomrule
\end{tabular}}
\end{subtable}
\vspace{0pt}
\end{table*}

To analyze the internal evolution of the latent diffusion process beyond image-level metrics, we examine second-order trajectory statistics during DDIM sampling.
In particular, we are interested in whether the predicted latent trajectory exhibits second-order evolution dynamics similar to those of the ground-truth target diffusion trajectory. Let $\{\mathbf{z}_t\}_{t=0}^{T}$ denote the sequence of latents generated along the sampling trajectory.
We define the discrete curvature at timestep $t$ as the second-order finite difference:
\begin{equation}
\kappa_t = \left\| \mathbf{z}_{t+1} - 2\mathbf{z}_t + \mathbf{z}_{t-1} \right\|_2.
\end{equation}
This quantity approximates the acceleration of the discrete trajectory and measures deviation from locally linear evolution in latent space. Under a well-aligned initialization, diffusion trajectories are expected to evolve smoothly under the pretrained score field. A large second-order deviation may therefore indicate mismatch with the learned diffusion dynamics. By comparing the distributions of $\kappa_t$ across methods using the 1-Wasserstein distance, we quantify differences in second-order trajectory statistics relative to the ground-truth target diffusion trajectory.

For each sample, we compute the curvature sequence for (i) the ground-truth (GT) target view sampling trajectory with target view initial latent ${\mathbf{z}}^{\mathrm{inv}}_{\mathrm{tar}}$, (ii) sampling trajectory with our Diffusion Dynamics Aligned Fused latent $\tilde{\mathbf{z}}^{\mathrm{inv}}_{\mathrm{tar}}$ (equation \ref{eq:final_latent}), and (iii) sampling trajectory with direct noise addition from reference image latent fused $\mathbf{z}_{\text{fused}}$ (equation \ref{eq:fused_latent}).
To quantify differences in second-order evolution statistics, we compute the Wasserstein distance between curvature distributions, $W(\kappa^{\text{GT}}, \kappa^{\text{fusion}})$, where lower values indicate closer similarity to the target diffusion trajectory.

\noindent \textbf{Full-Trajectory Curvature Dynamics.} We first compute the Wasserstein distance between full curvature sequences which is 30 DDIM Sampling steps for different frame-distance buckets.
Results are shown in Table~\ref{tab:curvature_full}. As the metrics suggest in Table \ref{tab:curvature_full}, diffusion sampling starting with our diffusion dynamics aligned fusion latent $\tilde{\mathbf{z}}^{\mathrm{inv}}_{\mathrm{tar}}$ consistently reduces curvature mismatch and is closer to the target sampling trajectory with the largest improvement observed in the medium viewpoint changes which is 10--20 frames in a 30-frame sweep. For small viewpoint shifts of 0-10 frames, the trajectory mismatch is less pronounced as the input and target views lie relatively close to each other on the diffusion manifold and using reference view noise directly in $\mathbf{z}_{\text{fused}}$ evolves relatively better during sampling compared to more extreme viewpoint changes.

\noindent \textbf{{Temporal Bin-wise Curvature Dynamics.}}
During inference, DDIM sampling proceeds from $t\!\approx\!601$ to $t\!=\!0$ in 30 steps
($\{601,581,\dots,21,0\}$).
To analyze phase-wise trajectory behavior, we partition the 30 steps into five equal bins
(6 steps each), corresponding to contiguous timestep intervals:
Bin1: $t=601\!\rightarrow\!501$,
Bin2: $t=481\!\rightarrow\!381$,
Bin3: $t=361\!\rightarrow\!261$,
Bin4: $t=241\!\rightarrow\!141$,
Bin5: $t=121\!\rightarrow\!0$.
For each bin, we compute the Wasserstein distance between curvature distributions of the predicted and target trajectories.
Firstly from Table \ref{tab:curvature_temporal}, we see that for all frame distance buckets, the most significant improvement occurs in early and mid diffusion stages Bins 1-3. In later bins, both methods converge to similar curvature statistics, suggesting that improvements primarily arise from better {trajectory shaping} rather than late-stage denoising adjustments.


\begin{figure*}[t]
    \centering
    \begin{subfigure}[t]{0.38\linewidth}
        \centering
        \includegraphics[
            width=\linewidth,
            height=0.28\textheight,
            keepaspectratio,
            trim=0 20pt 0 20pt,
            clip
        ]{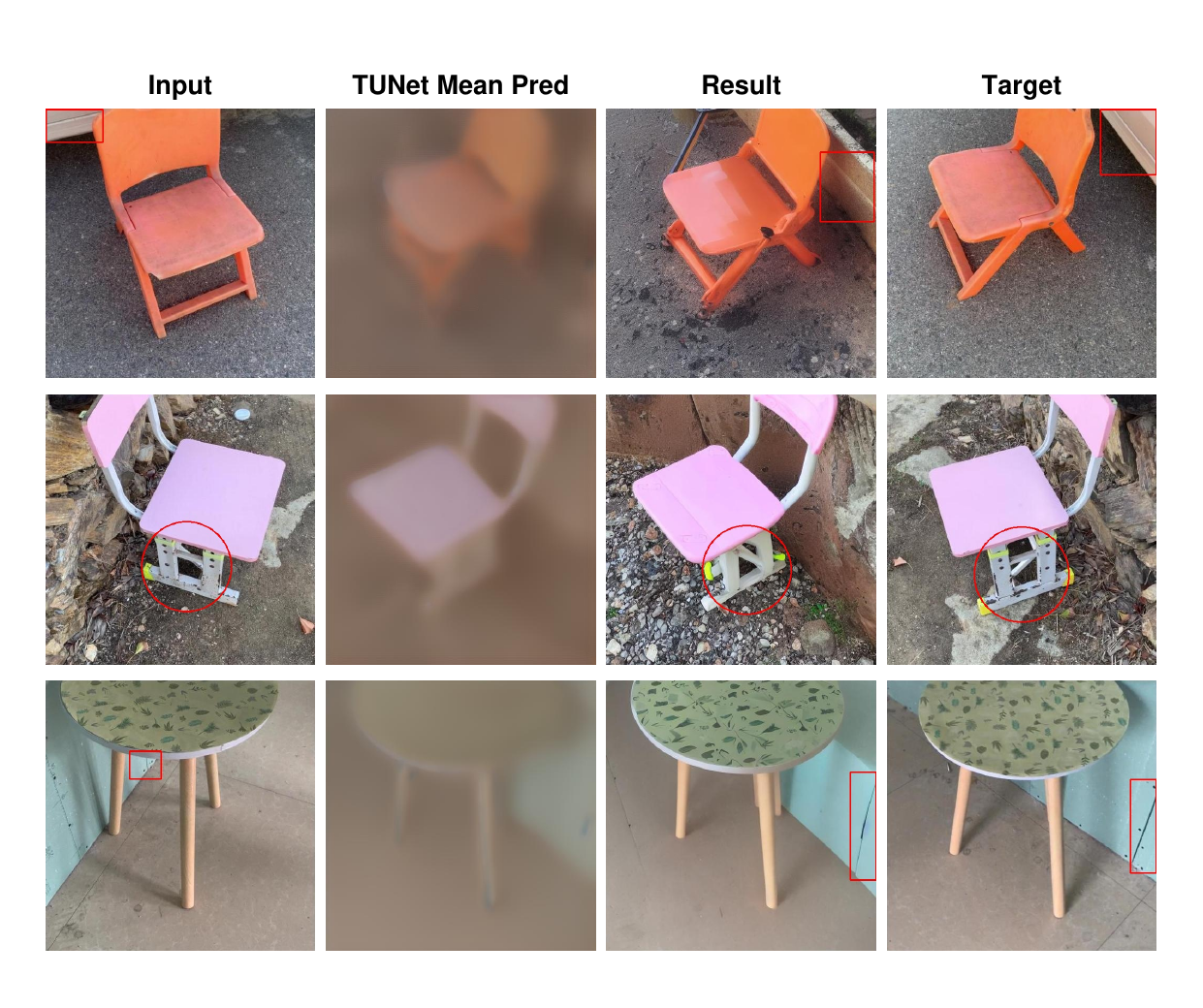}
        \caption{View-dependent details.}
        \label{fig:trajectory_and_geometry_a}
    \end{subfigure}\hfill
    \begin{subfigure}[t]{0.58\linewidth}
        \centering
        \includegraphics[
            width=\linewidth,
            height=0.28\textheight,
            keepaspectratio
        ]{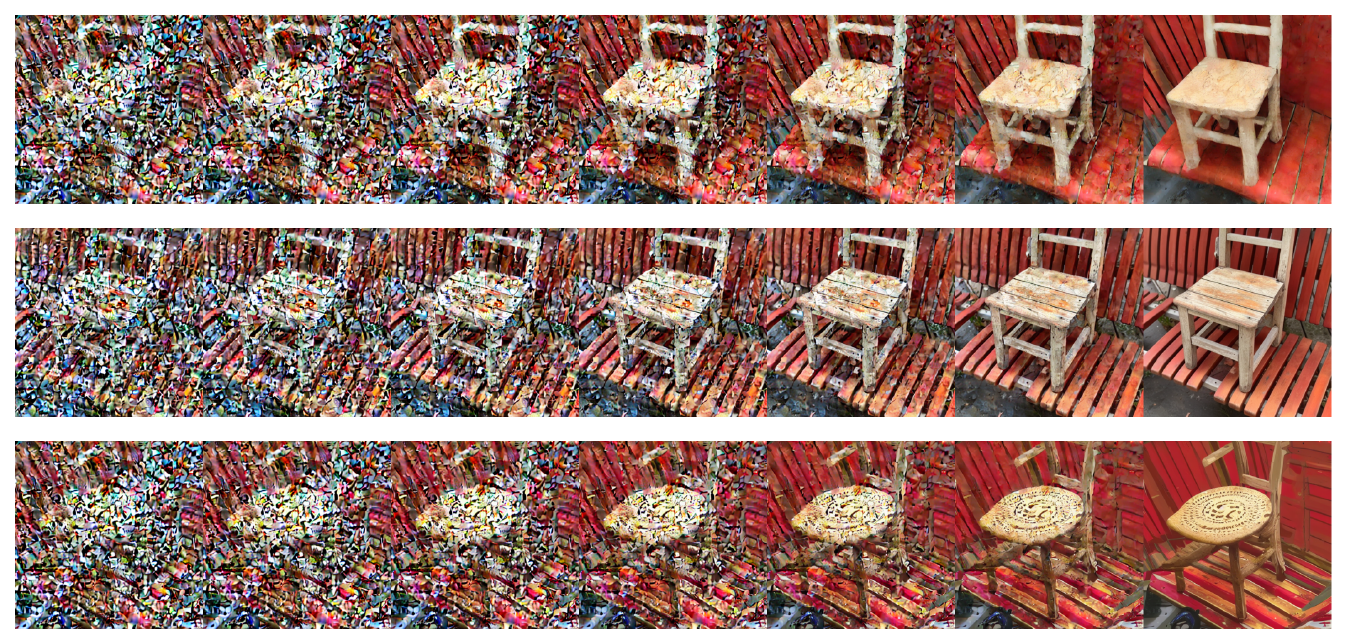}
        \caption{Sampling trajectories.}
        \label{fig:trajectory_and_geometry_b}
    \end{subfigure}

    \caption{
    \textbf{(a)} View-dependent geometric details absent in the mean prediction but re-synthesized in the final result.
    \textbf{(b)} Sampling trajectories under different initialization strategies. Row: (i) our diffusion-dynamics aligned latent $\tilde{\mathbf{z}}^{\mathrm{inv}}_{\mathrm{tar}}$,
    (ii) ground-truth target-view latent ${\mathbf{z}}^{\mathrm{inv}}_{\mathrm{tar}}$ (equation \ref{eq:final_latent}),
    and (iii) direct noise addition from the reference latent $\mathbf{z}_{\text{fused}}$ (equation \ref{eq:fused_latent}).
    }
    \label{fig:trajectory_and_geometry}
\end{figure*}

Next, we relate the results to the trajectory visualized in Figure \ref{fig:trajectory_and_geometry_b}. Specifically, we look at the first 4 sampling steps in the trajectory rows. In these diffusion sampling steps, our diffusion dynamics aligned fusion helps the sampling trajectory to stay close to the target trajectory at each of these steps. These diffusion steps are responsible for transforming the high variance Gaussian noise representations into target view aligned scene features. At these high noise regimes, our score corrected noise prediction $\boldsymbol{\epsilon}_\phi$ (equation \ref{eq:noise_prediction}), helps the evolution of $\tilde{\mathbf{z}}^{\mathrm{inv}}_{\mathrm{tar}}$ to stay closer to the target statistics. As the latents move to cleaner regimes, the view synthesised with $\tilde{\mathbf{z}}^{\mathrm{inv}}_{\mathrm{tar}}$ produces consistent target view geometry. In the 3rd row of the figure, we see that the trajectory initialised with $\mathbf{z}_{\text{fused}}$  deviates from the target trajectory producing a visually different chair as well as the geometry that is not aligned with the target view.
The results in Table \ref{tab:curvature_temporal} reflect the same phenomena where in the earlier Bins 1-4 our sampling started from $\tilde{\mathbf{z}}^{\mathrm{inv}}_{\mathrm{tar}}$ stays closer to the target trajectory which finally produces better results. The Wasserstein distance statistics have lesser difference in the later bins, Bin 4 and Bin 5, as at these points the trajectory starts converging to the clean data manifold.


\noindent \textbf{View Dependent Synthesis}: Figure \ref{fig:trajectory_and_geometry_a} illustrates a consistent phenomenon across multiple examples.
The TUNet mean prediction successfully translates coarse structure and global pose to the target viewpoint. Object silhouettes and primary geometry are preserved.
However, fine-scale, view-dependent details are systematically absent in the mean prediction. These include thin linear structures, secondary contours, and background elements whose appearance changes under camera transformation. We can see from the results that most of the scene details are absent in the TUNet mean predictions and are re-synthesised after the fusion step with the diffusion priors.
After diffusion dynamics aligned fusion, these missing structures are re-synthesized in a manner consistent with the target geometry.
In the first row, the horizontal line appearing on the structure behind the orange chair is synthesised in  the correct direction according to the view transformation. In the second  row, the proper gap and structure of the leg of the pink chair appears in the result from an extreme view transformation. In the third row, the vertical line on the background wall is synthesised consistent with the target geometry. In all of these results, the TUNet mean prediction does not capture these parts of the scene and these are implicitly view transformed in a geometric consistent manner with the diffusion priors.


\begin{table*}[t]
\centering
\caption{(a) Comparison for 3 classes - chairs, sofa, tables. Resolution is $256\times256$. (b) Comparison for 167 classes. Resolution is $90\times90$.}
\vspace{0pt}
\begin{subtable}[t]{0.48\textwidth}
  \centering
  \caption{}
  \vspace{0pt}
  \label{tab:comp_256}
  \begin{tabular}{@{}lccc@{}}
    \toprule
    Method & LPIPS $\downarrow$ & PSNR $\uparrow$ & SSIM $\uparrow$ \\
    \midrule
    GIBR & $0.510$ & $\textbf{17.61}$ & $\textbf{0.554}$ \\
    Ours & $\textbf{0.490}$ & $15.71$ & $0.523$ \\
    \bottomrule
  \end{tabular}
\end{subtable}\hfill
\begin{subtable}[t]{0.48\textwidth}
  \centering
  \caption{}
  \vspace{0pt}
  \label{tab:comp_90}
  \begin{tabular}{@{}lcccc@{}}
    \toprule
    Method & LPIPS $\downarrow$ & PSNR $\uparrow$ & SSIM $\uparrow$ & FID $\downarrow$ \\
    \midrule
    NViST & $0.448$ & $14.31$ & $0.566$ & $91.63$ \\
    Ours & $\textbf{0.409}$ & $\textbf{16.16}$ & $\textbf{0.578}$ & $\textbf{65.50}$ \\
    \bottomrule
  \end{tabular}
\end{subtable}
\vspace{0pt}
\end{table*}

\subsection{Quantitative Comparison}\vspace{0pt}

\textbf{3-class model:} Comparison with GIBR on 3 classes at a resolution of $256\times256$ is shown in Table \ref{tab:comp_256}. Input and target pairs from 168 unseen scenes are used for testing. We outperform GIBR in terms of LPIPS. GIBR trains the entire diffusion process in the RGB space and also uses multiple views while training and volume rendering to generate the final image. Thus, GIBR does better in terms of PSNR and SSIM. However, training diffusion in pixel space is very expensive. On the other hand, we only train our TUNet with 148M parameters using latents.

\noindent \textbf{167-class model:} Comparison with NViST at a resolution of $90\times90$ is shown in Table \ref{tab:comp_90}. Input and target pairs from 360 unseen scenes are used for testing. Here, we see that our method performs better in terms of LPIPS, PSNR, SSIM, and FID. We show the synthesized results in Figure \ref{fig:grid_2}. In the case of the kettle, we can see that the unobserved region is synthesized with high fidelity. Similarly, in the case of a bowl (third row, middle column), the shadow is faithfully synthesized.




\subsection{Qualitative Comparison}

\begin{figure*}[!tp]
    \centering
    \begin{subfigure}[t]{0.29\textwidth}
        \centering
        \includegraphics[width=1.05\linewidth]{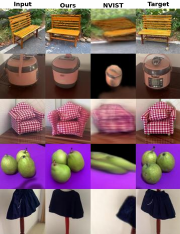}
        \caption{}
        \label{fig:comparison_and_multiview_a}
    \end{subfigure}\hfill
    \begin{subfigure}[t]{0.66\textwidth}
        \centering
        \includegraphics[
            width=\linewidth,
            height=0.35\textheight,
            keepaspectratio
        ]{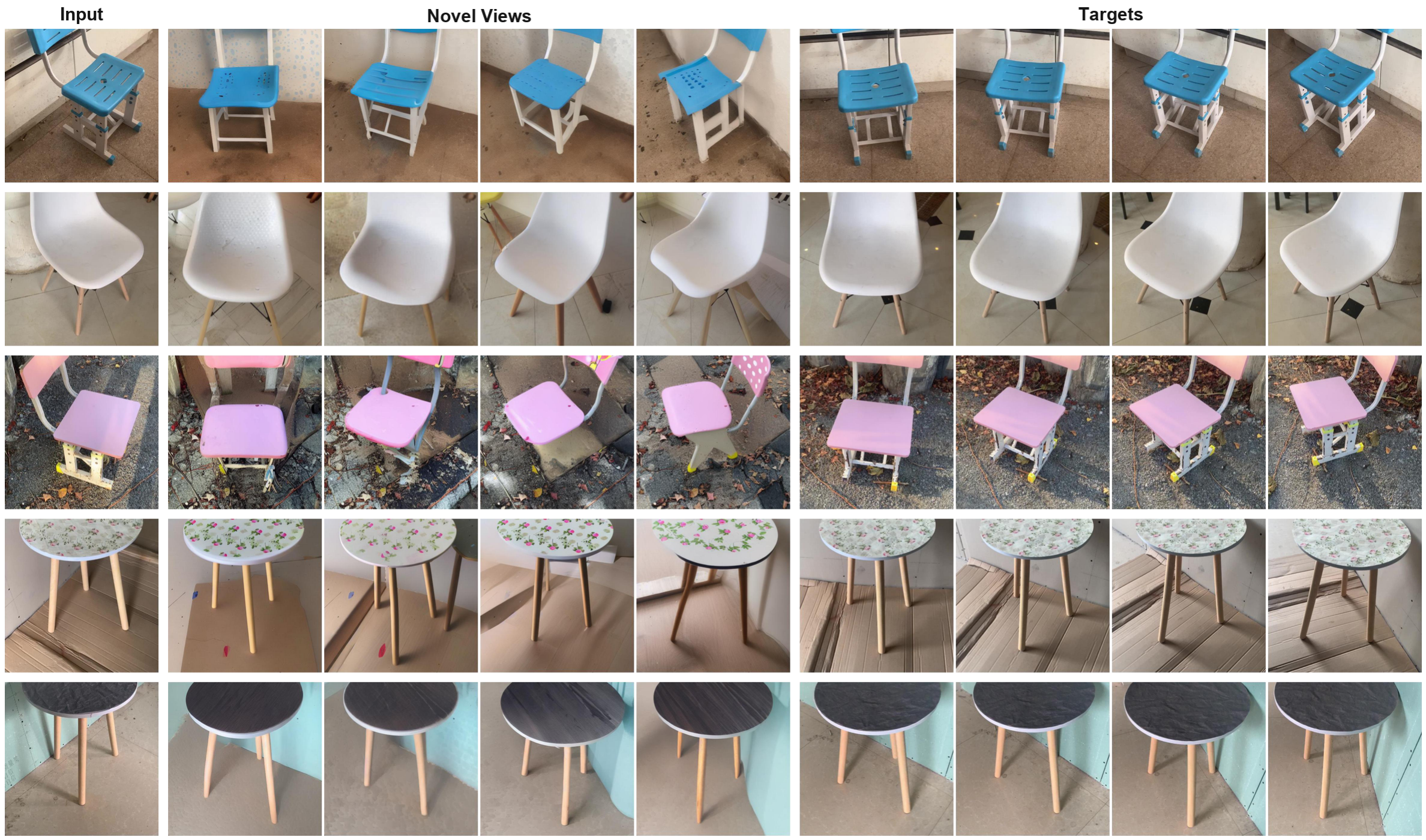}
        \caption{}
        \label{fig:comparison_and_multiview_b}
    \end{subfigure}

    \caption{
    \textbf{(a)} Comparison with NViST on unseen test scenes from five classes.
    \textbf{(b)} Multi-view consistency across a 30-frame sweep from a single input image.
    }
    \label{fig:comparison_and_multiview}
\end{figure*}

We compare our results with NViST \cite{jang2024nvistwildnewview} in Figure \ref{fig:comparison_and_multiview_a}. It is evident that our method synthesizes the target with high fidelity and is able to generate results for near as well as far target views, where NViST fails.

We also show multi-view generation using single frame in Figure \ref{fig:comparison_and_multiview_b}. We use one of the first five frames as the input frame and generate 16$^{th}$, 19$^{th}$, 22$^{nd}$, and 25$^{th}$ frame. Compared to the respective target, we observe that our method generates faithful output with consistent geometry.

\begin{figure*}[!tp]
  \centering
  \begin{subfigure}[t]{0.65\textwidth}
    \centering
    \includegraphics[width=\linewidth]{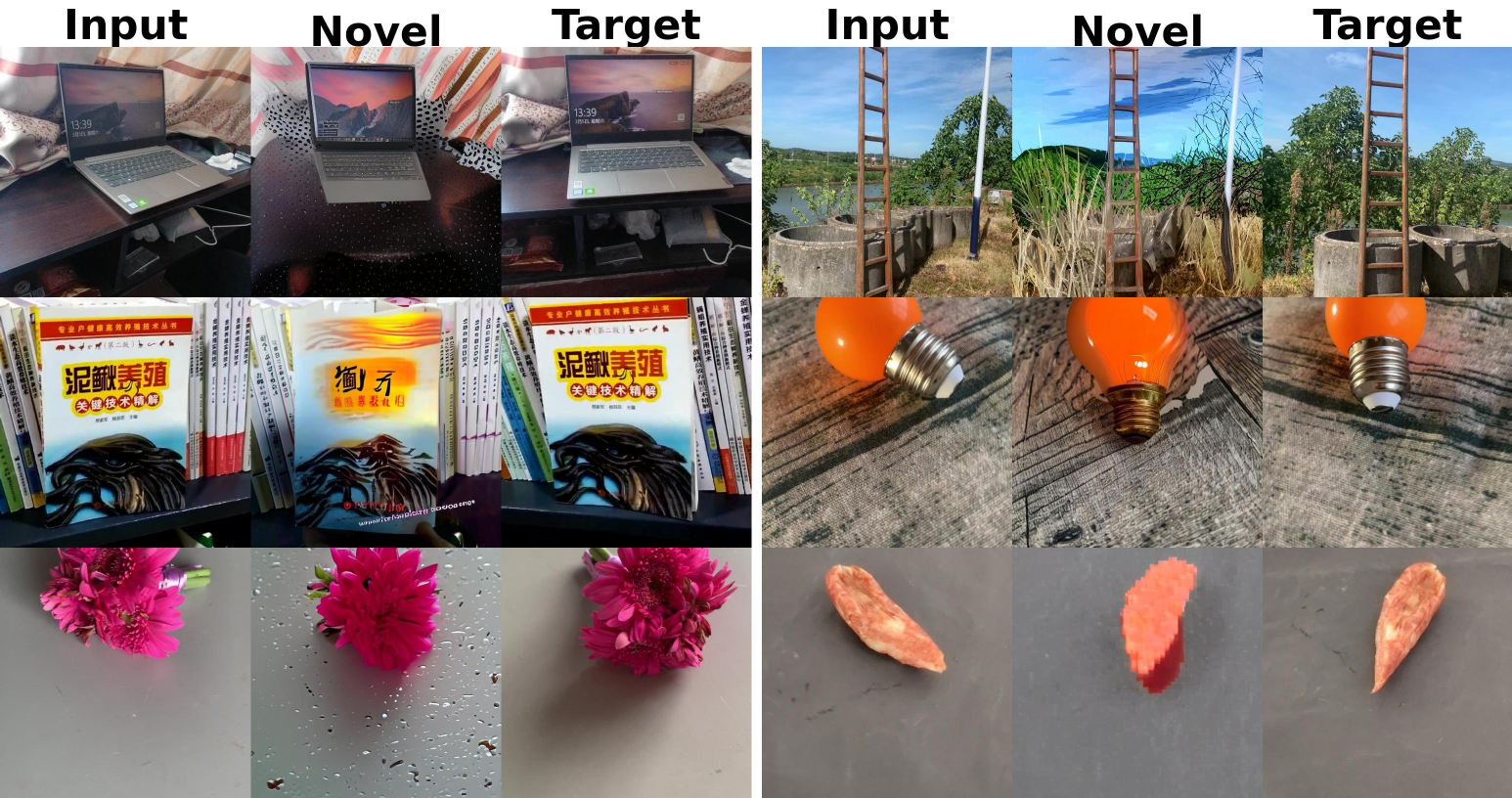}
    \caption{Results on 6 unseen classes from MVImgNet}
    \label{fig:unseen_classes}
  \end{subfigure}\hfill
  \begin{subfigure}[t]{0.33\textwidth}
    \centering
    \includegraphics[width=\linewidth]{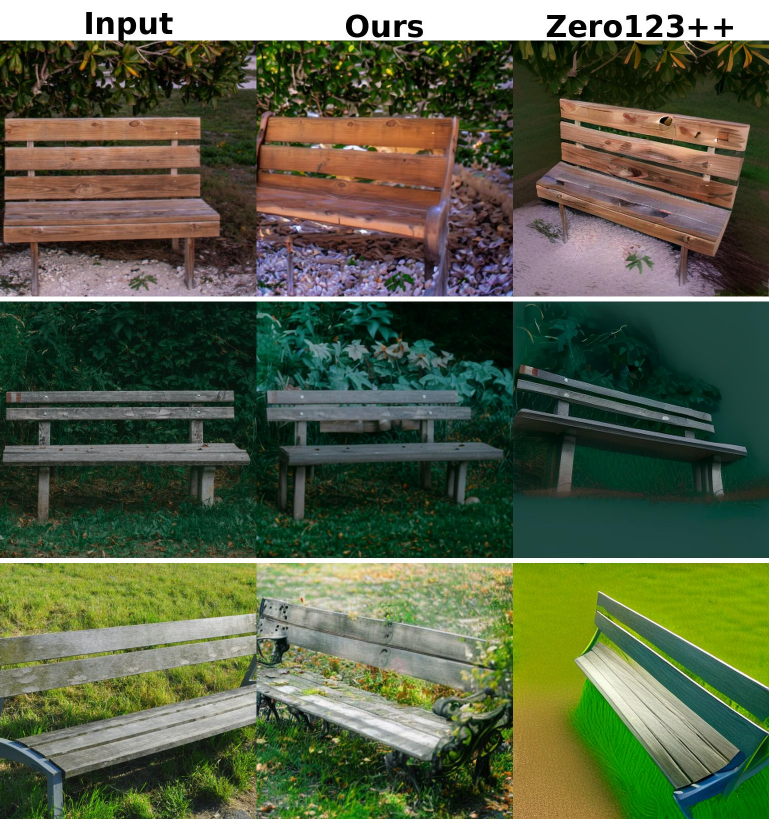}
    \caption{Out-of-domain images}
    \label{fig:ood_results}
  \end{subfigure}
  \caption{Results on unseen classes of MVImgNet and OOD images}
\end{figure*}

\noindent \textbf{Unseen Classes}: We show qualitative results on 6 unseen classes in Figure \ref{fig:unseen_classes}. We cover outdoor and indoor scenes, as well as include large and small object classes in the test set. Even on unseen classes, we are able to generate high-resolution reconstruction for a diverse set of scenes.

\noindent \textbf{Out of Domain Data}: To evaluate zero‑shot generalization beyond MVImgNet, we assembled an out‑of‑domain test set by downloading freely‑licensed photographs from Unsplash \cite{unsplash2025} featuring natural scenes.
We show the results and compare with Zero123++ \cite{shi2023zero123singleimageconsistent} in Figure \ref{fig:ood_results}. While both methods successfully generate plausible novel viewpoints, our approach produces more faithful surface textures and preserves natural scene characteristics.

\begin{table*}[!t]
\centering
\vspace{0pt}
\caption{(a) Ablation study using 3-class model. (b) Analysis of fusion strategies}
\vspace{0pt}
\begin{subtable}[t]{0.49\textwidth}
  \centering
  \caption{}
  \vspace{0pt}
  \label{tab:ablation}
  \resizebox{.98\textwidth}{!}{%
  \begin{tabular}{@{}lccc@{}}
    \toprule
    Setting & LPIPS $\downarrow$ & PSNR $\uparrow$ & SSIM $\uparrow$ \\
    \midrule
    Concat (class, cam. w inp.) & {0.508} & {15.38} & {0.516} \\
    w/o cross-attention & {0.506} & {15.41} & {0.515} \\
    \textbf{Full model} & \textbf{0.492} & \textbf{15.71} & \textbf{0.523} \\
    \bottomrule
  \end{tabular}}
\end{subtable}\hfill
\begin{subtable}[t]{0.49\textwidth}
  \centering
  \caption{}
  \vspace{0pt}
  \label{tab:perceptual_ablation}
  \resizebox{.98\textwidth}{!}{%
  \begin{tabular}{@{}lcccc@{}}
    \toprule
    Initial latent & LPIPS $\downarrow$ & PSNR $\uparrow$ & SSIM $\uparrow$ & FID $\downarrow$ \\
    \midrule
    $\tilde{\mathbf{z}}^{\mathrm{inv}}_{\mathrm{tar}}$ & 0.495 & 15.41 & 0.521 & 69.86 \\
    $\mathbf{z}_{\text{fused}}$ & 0.521 & 14.76 & 0.457 & 102.72 \\
    \bottomrule
  \end{tabular}}
\end{subtable}
\vspace{0pt}
\end{table*}

\subsection{Ablation Study}
\label{sec:ablation}
\textbf{Architecture Design}: 
The model design ablation results are presented in Table \ref{tab:ablation}. We evaluate the following settings. In the first setting, {Concat}, we concatenate class and camera embeddings with the input, but do not inject them into every ResNet block. Here, we see that there is a significant performance drop. Second, {w/o cross-attn}, where we remove all cross-attention layers. The results degrade for all three metrics.

\noindent \textbf{Fusion Comparison}: 
We compare perceptual and image quality assessment metrics for images generated with our diffusion dynamics aligned fusion latent $\tilde{\mathbf{z}}^{\mathrm{inv}}_{\mathrm{tar}}$ and direct noise added fusion latent $\mathbf{z}_{\text{fused}}$ in Table \ref{tab:perceptual_ablation}. Images synthesized with $\tilde{\mathbf{z}}^{\mathrm{inv}}_{\mathrm{tar}}$ achieves better results in all metrics.

Our curvature analysis provides mechanistic insight where improved image quality correlates with reduced second-order structural deviation in latent space. The curvature measures alignment with the target diffusion trajectory. 
The observed improvements therefore indicate that diffusion dynamics aligned fusion better preserves structural geometry during novel-view synthesis.

\begin{table}[t]
    \centering
    \caption{Effect of diffusion timestep $t$.}
    \label{tab:timestep_ablation}

    \small
    \setlength{\tabcolsep}{4pt}
    \renewcommand{\arraystretch}{1.0}

    \begin{tabular}{@{}lcccc@{}}
        \toprule
        $t$ & LPIPS $\downarrow$ & PSNR $\uparrow$
            & SSIM $\uparrow$ & FID $\downarrow$ \\
        \midrule
        400 & 0.510 & 15.37 & 0.528 & 150.28 \\
        600 & 0.491 & 15.41 & 0.522 & 93.57  \\
        800 & 0.550 & 15.01 & 0.502 & 197.13 \\
        \bottomrule
    \end{tabular}
\end{table}

\noindent \textbf{Different Diffusion Timesteps}:
 We compare the performance at three different timesteps $t=  400, 600 ,800$. As shown in Table \ref{tab:timestep_ablation}, decreasing the diffusion timestep from $t=600$ to $t=400$ leads to a degraded performance as reflected in the scores. 
 In our experiments, we observe that at $t=400$, the loss is very high compared to the case of $t=600$. 
 This indicates that training is harder for a single-step translation with TUnet for $t=400$. 
 At $t=400$, the noise level and the inversion trajectory are insufficient for meaningful variance-based fusion. The latent still retains the dominant low-frequency structure, reducing the fusion strategy's ability to recover and refine high-frequency content. The performance is worst at $t=800$ as the inversion at this timestep loses most of its signal to perform any effective translation. 
In contrast, $t=600$ allows for effective view translation aided by noise fusion to recover better quality results upon sampling.

\noindent \textbf{Different Stable Diffusion Pipelines}: 
We evaluate our pipeline using three diffusion backbones. Stable Diffusion v1.5 \cite{Rombach_2022_CVPR} is the pipeline we use by default in all of our experiments. Furthermore, we compare our default pipeline with v2.1, and Dreamlike Photoreal 2.0. 
The three pipelines differ in architecture, training data, and learned denoising priors.
v1.5 employs the original UNet and OpenAI CLIP ViT-L/14 \cite{radford2021learning} trained on LAION-2B \cite{schuhmann2022laion} whereas v2.1 introduces an upgraded UNet with OpenCLIP ViT-H/14 trained on the higher-quality LAION-5B dataset.
Dreamlike Photoreal 2.0 is a domain-specific fine-tuned version of v2.1 emphasizing photorealistic imagery.
As shown in Table~\ref{tab:sd_pipeline_ablation}, performance remains consistent across models, indicating robustness of our framework.
v2.1 achieves slightly lower FID, reflecting improved generative realism, while v1.5 attains marginally higher PSNR/SSIM. Dreamlike Photoreal 2.0 shows increased artifacts, likely due to its photoreal art domain fine-tuning, which makes it perform worse for natural images.
\begin{table}[h]
  \centering
  \caption{Comparison across different SD pipelines.}
  \label{tab:sd_pipeline_ablation}
  \resizebox{\linewidth}{!}{%
  \begin{tabular}{@{}lcccc@{}}
    \toprule
     Pipeline & LPIPS $\downarrow$ & PSNR $\uparrow$ & SSIM $\uparrow$ & FID $\downarrow$ \\
    \midrule
    Stable Diffusion v1.5 & {0.491} & {15.41} & {0.522} & {93.57} \\
    Stable Diffusion 2.1      & {0.491} & {15.27} & {0.522} & {88.95} \\
    {Dreamlike Photoreal 2.0} & 0.507  & 15.15  & 0.503  & 101.84  \\
    \bottomrule
  \end{tabular}
  }
\end{table}

\noindent \textbf{Frame Distance Analysis.}
To analyze how viewpoint change affects reconstruction quality, we group test pairs based on the temporal distance between the source frame and the target frame. 
Each source frame in the test set is paired with all valid target frames to synthesize novel views, resulting in a test set of 18,358 source–target pairs. 
We divide these pairs into buckets based on frame distance and compute the mean and standard deviation of VGG LPIPS, PSNR, and SSIM within each bucket.

As shown in Table~\ref{tab:frame_distance_analysis}, performance gradually degrades as the frame distance increases, reflecting the increasing difficulty of synthesizing views with larger geometric transformations and occlusions. 
Nevertheless, the degradation remains gradual, indicating that the proposed framework maintains stable performance even under long-range viewpoint changes.

\begin{table}[h]
\centering
\caption{Performance as a function of source--target frame distance. 
Metrics are reported as mean $\pm$ standard deviation across all test pairs in each bucket.}
\label{tab:frame_distance_analysis}
\resizebox{\linewidth}{!}{
\begin{tabular}{lccccc}
\toprule
Frame Distance & $N$ & LPIPS $\downarrow$ & PSNR $\uparrow$ & SSIM $\uparrow$ \\
\midrule
0--5   & 196  & $0.4741 \pm 0.0455$ & $16.28 \pm 1.92$ & $0.543 \pm 0.19$ \\
6--10  & 3375 & $0.4863 \pm 0.0483$ & $15.82 \pm 2.01$ & $0.528 \pm 0.17$ \\
11--16 & 8811 & $0.4914 \pm 0.0499$ & $15.63 \pm 2.06$ & $0.525 \pm 0.17$ \\
17--21 & 4980 & $0.4949 \pm 0.0508$ & $15.55 \pm 2.11$ & $0.517 \pm 0.17$\\
21--25 & 996  & $0.4996 \pm 0.0517$ & $15.44 \pm 2.14$ & $0.513 \pm 0.17$ \\
\bottomrule
\end{tabular}
}
\end{table}

\subsection{RealEstate10K}
\begin{table}[!t]
\vspace{0pt}
\centering
\caption{Results on 1K pairs of RealEstate10K. Images are uniformly sampled at random from different scenes.}
\label{tab:mid_long_lpips_psnr_ssim}
\resizebox{\linewidth}{!}{%
\begin{tabular}{lcccccc}
\toprule
& \multicolumn{3}{c}{Mid-range (30--60 frames)} & \multicolumn{3}{c}{Long-range (60--120 frames)} \\
\cmidrule(lr){2-4}\cmidrule(lr){5-7}
Method & LPIPS $\downarrow$ & PSNR $\uparrow$ & SSIM $\uparrow$
       & LPIPS $\downarrow$ & PSNR $\uparrow$ & SSIM $\uparrow$ \\
\midrule
VIVID & 0.523 & 13.83 & 0.439 & 0.594 & 12.69 & 0.410 \\
Ours  & 0.503 & 15.04 & 0.479 & 0.609 & 13.44 & 0.448 \\
\bottomrule
\end{tabular}}
\vspace{0pt}
\end{table}
We compare our results with VIVID \cite{elata2025novel} that shows strong performance. We use 1K images for testing. In Table \ref{tab:mid_long_lpips_psnr_ssim}, we can see that our method performs better in terms of LPIPS, PSNR, and SSIM, except for long range LPIPS.

\begin{figure}
\centering
\includegraphics[width=0.9\columnwidth]{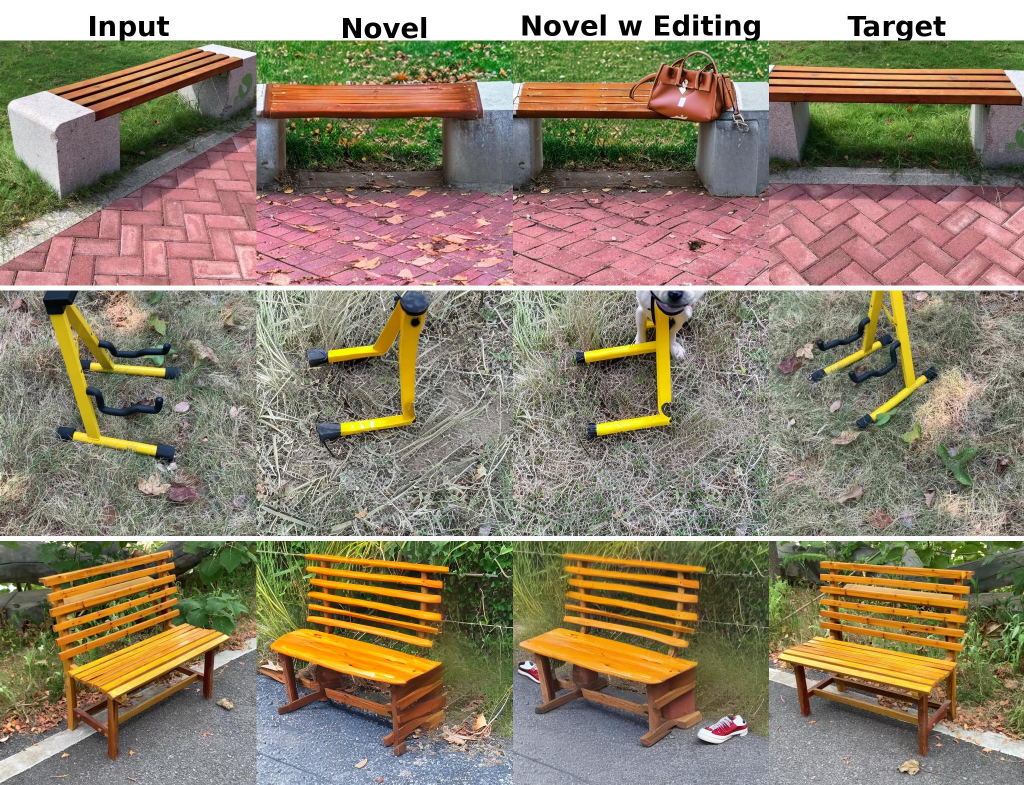}
\caption{We use textual prompts to edit the novel view reconstructions as desired. In the first row, we place a bag on the bench. In the second row, a dog is introduced in the scene. In the third row, a pair of shoes are placed by the bench.}
\label{fig:editing_figure}
\end{figure}
\subsection{Novel View Synthesis and Editing}
As we work with DDIM inverted noise, we have the flexibility to extend our pipeline to editing via textual prompts during the final sampling. For our novel view reconstruction, we use BLIP 
\cite{li2022blipbootstrappinglanguageimagepretraining_} generated captions for the input image $\mathbf{x}_{ref}$. We edit these prompts to add desired scenarios in our reconstructed novel view. Figure \ref{fig:editing_figure} shows results of our editing mechanism.

\section{Discussion, Limitations and Future Work}

Our formulation treats single-image novel-view synthesis as the construction
of a view-conditioned initialization for a frozen diffusion model. This
perspective clarifies both the division of responsibility between TUNet and
the pretrained diffusion prior, as well as the distinction between synthesizing a plausible
target-view image and explicitly reconstructing a scene for rendering.

\textbf{Viewpoint control through trajectory initialization:}
TUNet predicts a target-view structural mean from the source representation
and camera parameters. The fusion mechanism then constructs the noisy state
from which reverse DDIM sampling begins. Thus, the learned component controls
the requested view through the initial state, rather than by learning a new
camera-conditioned denoising field. The subsequent diffusion steps refine a
target-view hypothesis. These denoising steps should not be interpreted as a physical camera
trajectory or a sequence of explicitly rendered intermediate viewpoints.
Under fixed sampling conditions, target-camera information reaches the frozen
generator through the translated mean and the resulting score corrected initialization.
Consequently, the prior can complete and refine the proposed structure, but
it cannot be assumed to correct each camera information that is absent or incorrectly
represented in the mean prediction. For example, in Figure  \ref{fig:trajectory_and_geometry_a}, we cannot guarantee which of the target view features absent in the TUNet mean prediction will be re-synthesized correctly and which features will be missed or rendered incorrectly. The mechanism is implicit in its current form. 

\textbf{Source-conditioned refinement rather than texture copying:}
The source residual $\mathbf{z}^{\text{inv}}_{\text{ref},\sigma}$ in Equation ~\ref{eq:fused_latent} is a sample-specific
component of the source inversion state, not an explicitly aligned
target-view texture field. Its role is to influence the proposal $\mathbf{z}_{\mathrm{fused}}$ evaluated
by the frozen noise predictor towards the source scene. Equation~\ref{eq:final_latent} then uses the
predicted noise, rather than the unchanged source residual, in the final
sampling initialization. The correction is an
empirically motivated use of the pretrained denoiser, it is not a guaranteed
projection onto a target-view manifold.

\textbf{Scope and geometric consistency:}
Our method uses camera and ray-aware
conditioning, but does not maintain a shared 3D representation, explicitly
reason about visibility, or enforce that separately generated views are
projections of one consistent scene. The output is therefore a
camera-conditioned approximation of the target observation, with
unobserved content completed using the learned translation and the pretrained prior.

\textbf{Future Work:} Explicit correspondence, visibility-aware constraints, and
geometry-aware residual warping/transport are natural directions for strengthening
consistency while retaining the frozen-backbone formulation. Such extensions
would need to distinguish information supported by the source view from
content that must be inferred, and to assess which source features are
retained or altered by the score based residual correction.

\begin{figure}[H]
    \centering
    \setlength{\tabcolsep}{1pt}
    \renewcommand{\arraystretch}{1.0}

    \resizebox{\columnwidth}{!}{%
        \begin{tabular}{@{}c*{8}{c}@{}}
            \textbf{Input} &
            \includegraphics[width=1\columnwidth]{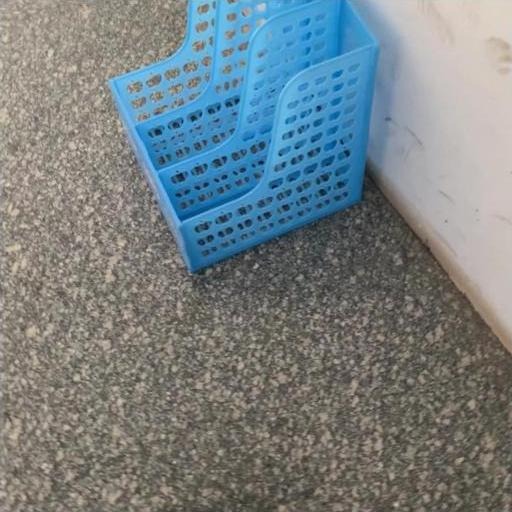} &
            \includegraphics[width=1\columnwidth]{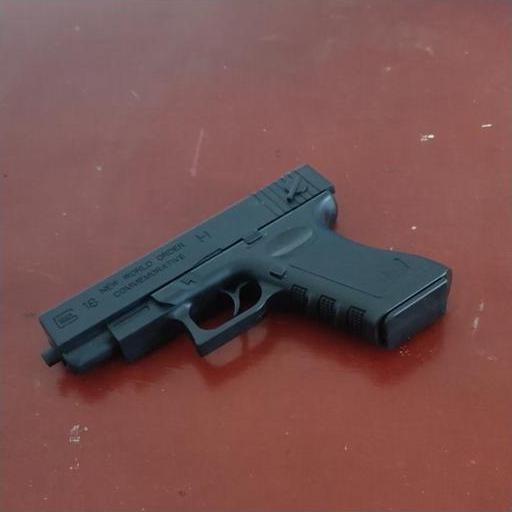} &
            \includegraphics[width=1\columnwidth]{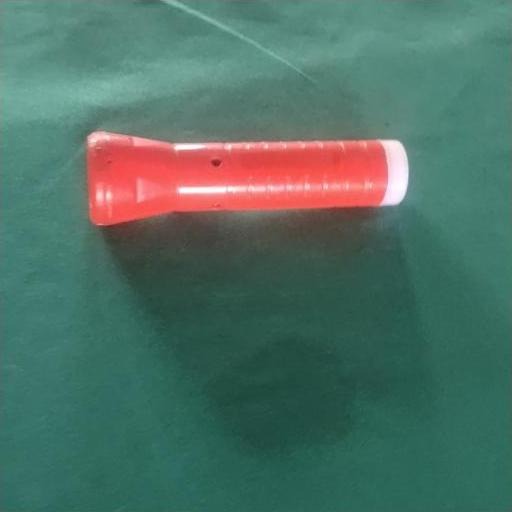} &
            \includegraphics[width=1\columnwidth]{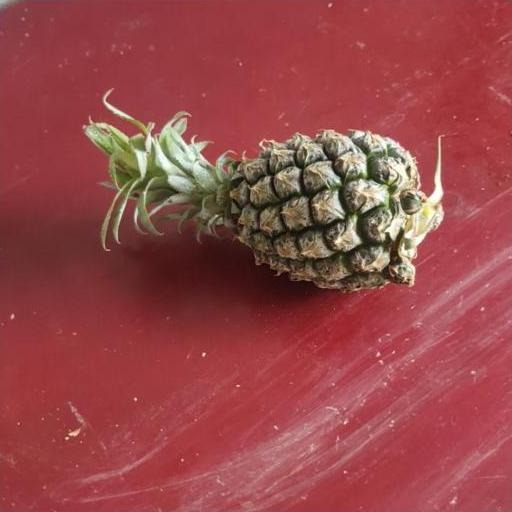} &
            \includegraphics[width=1\columnwidth]{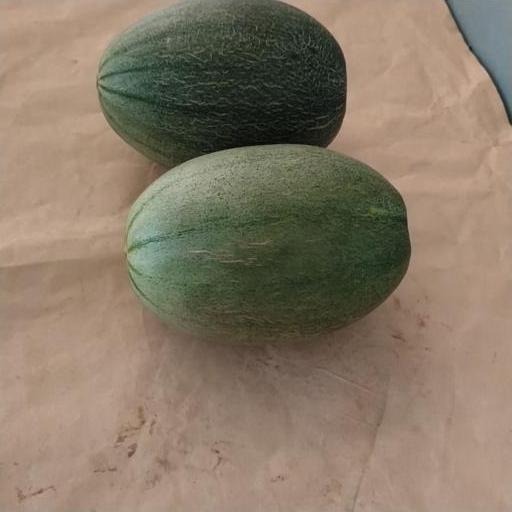} &
            \includegraphics[width=1\columnwidth]{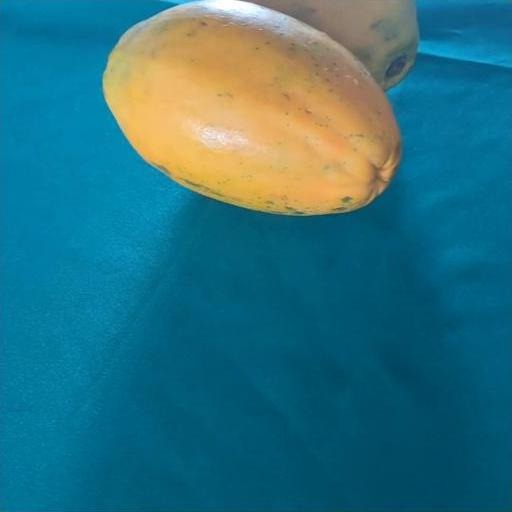} &
            \includegraphics[width=1\columnwidth]{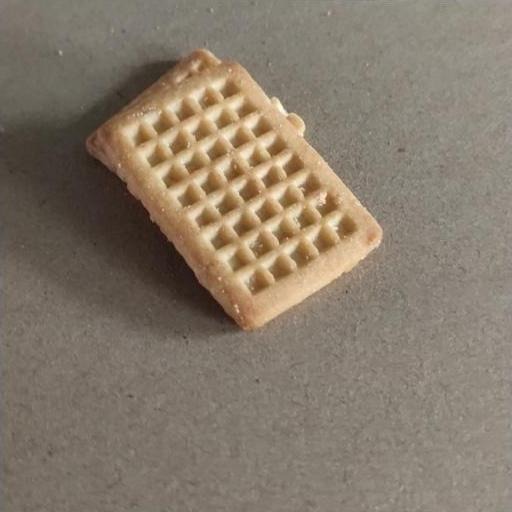} &
            \includegraphics[width=1\columnwidth]{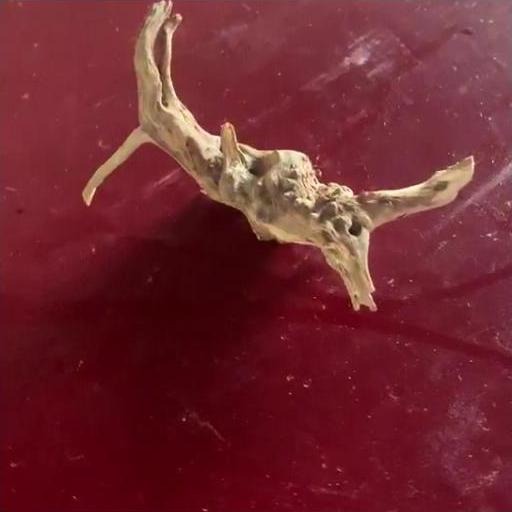} \\

            \textbf{Novel} &
            \includegraphics[width=1\columnwidth]{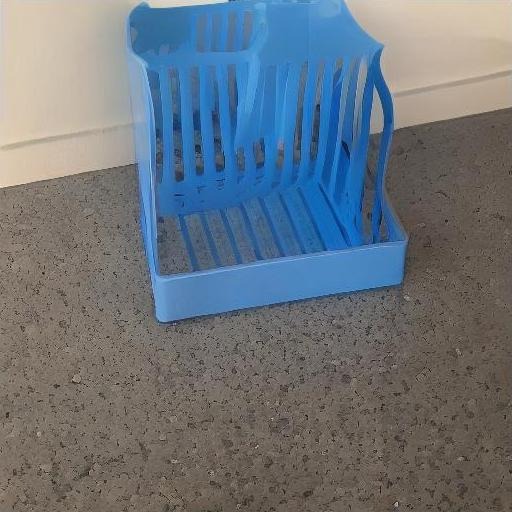} &
            \includegraphics[width=1\columnwidth]{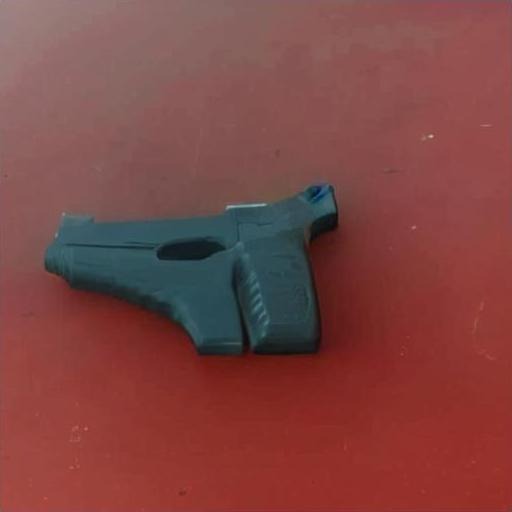} &
            \includegraphics[width=1\columnwidth]{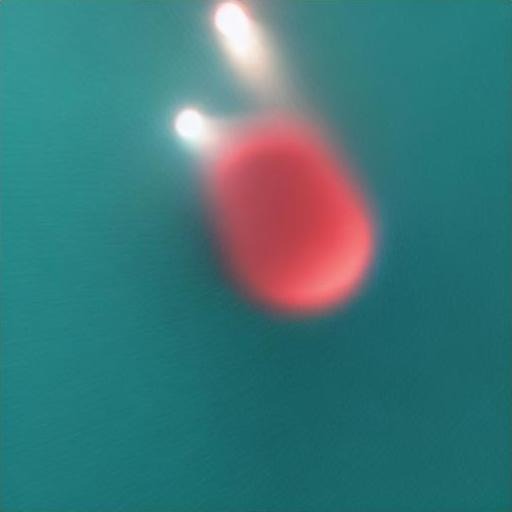} &
            \includegraphics[width=1\columnwidth]{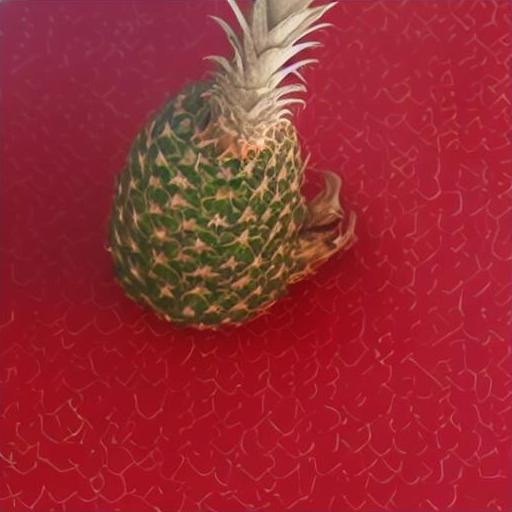} &
            \includegraphics[width=1\columnwidth]{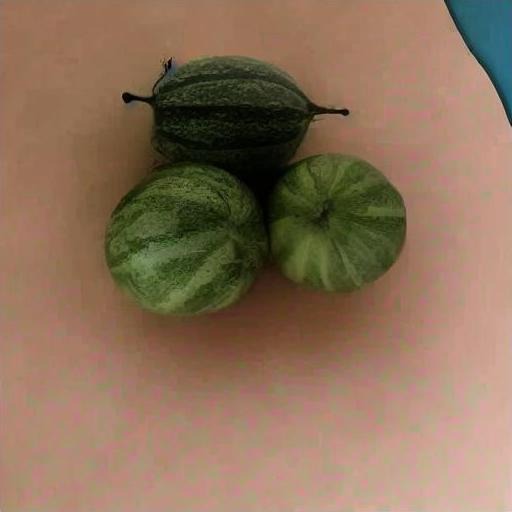} &
            \includegraphics[width=1\columnwidth]{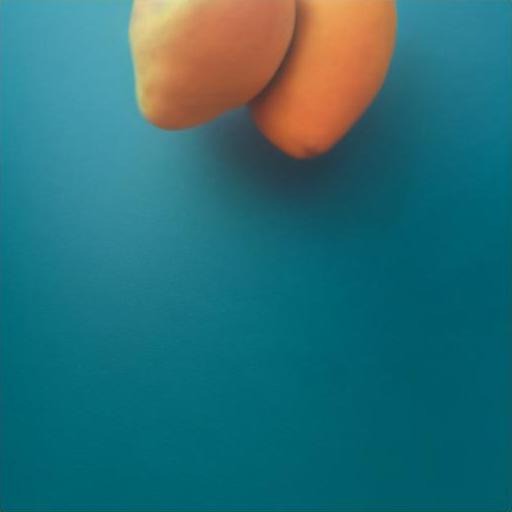} &
            \includegraphics[width=1\columnwidth]{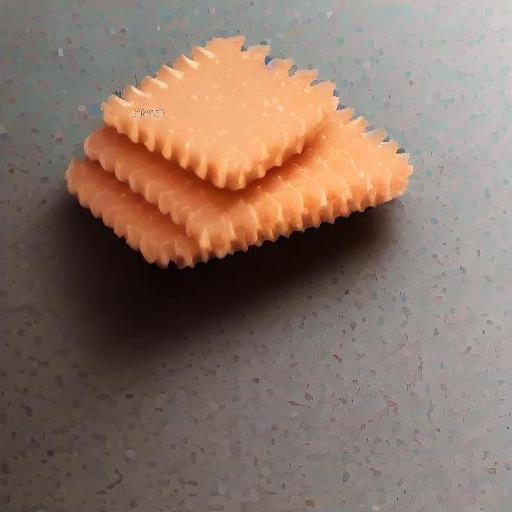} &
            \includegraphics[width=1\columnwidth]{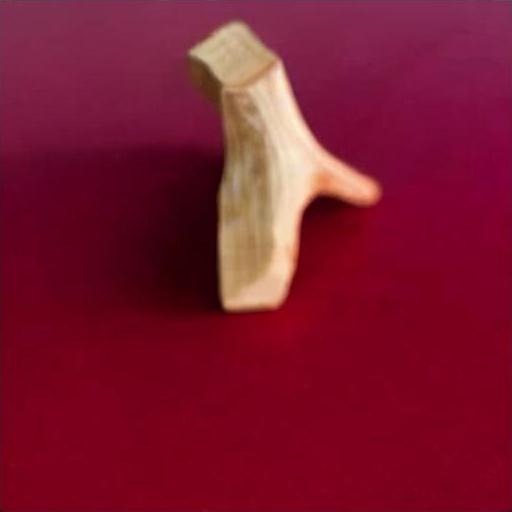} \\

            \textbf{Target} &
            \includegraphics[width=1\columnwidth]{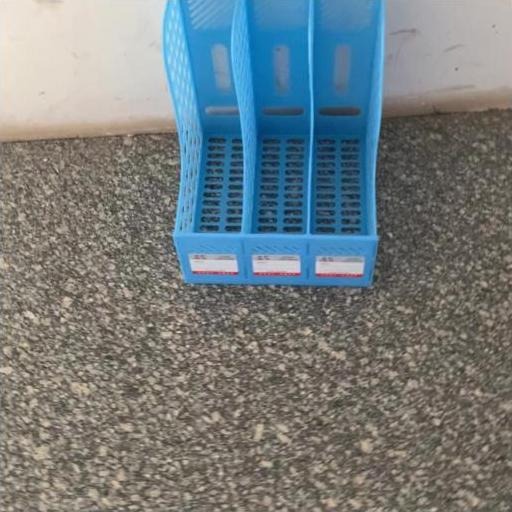} &
            \includegraphics[width=1\columnwidth]{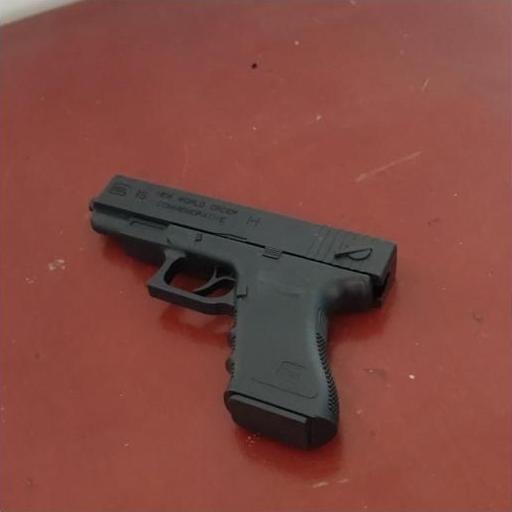} &
            \includegraphics[width=1\columnwidth]{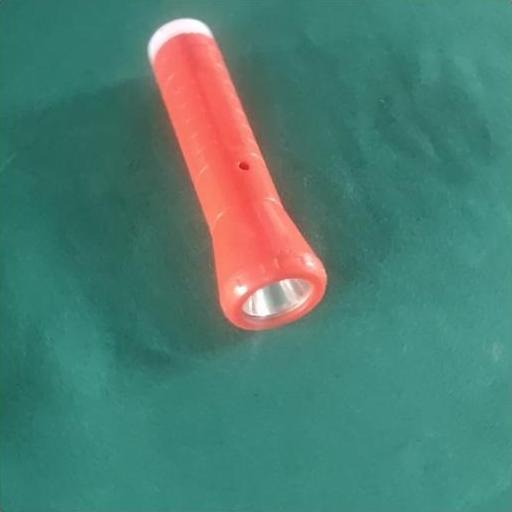} &
            \includegraphics[width=1\columnwidth]{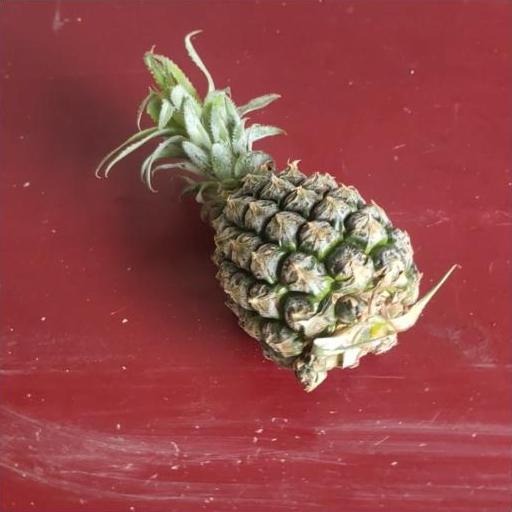} &
            \includegraphics[width=1\columnwidth]{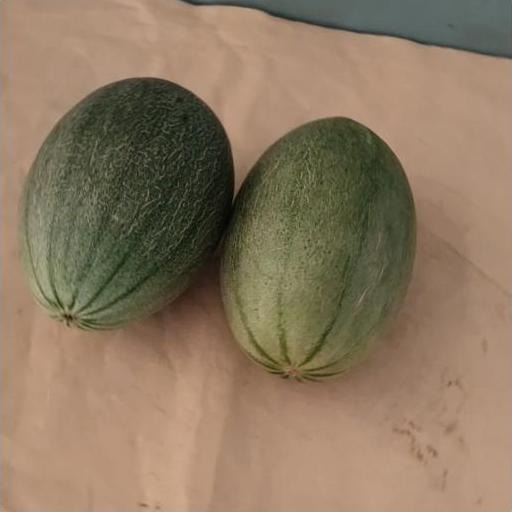} &
            \includegraphics[width=1\columnwidth]{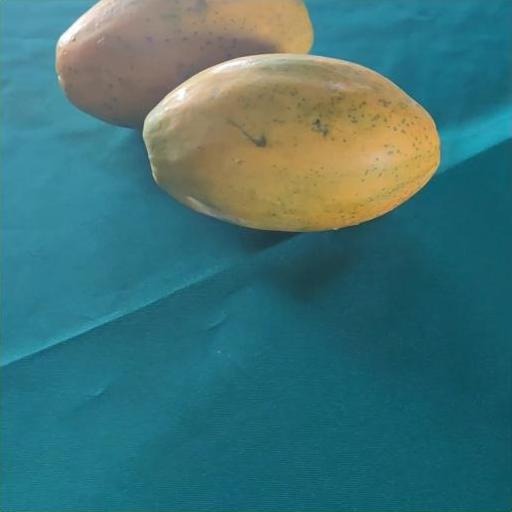} &
            \includegraphics[width=1\columnwidth]{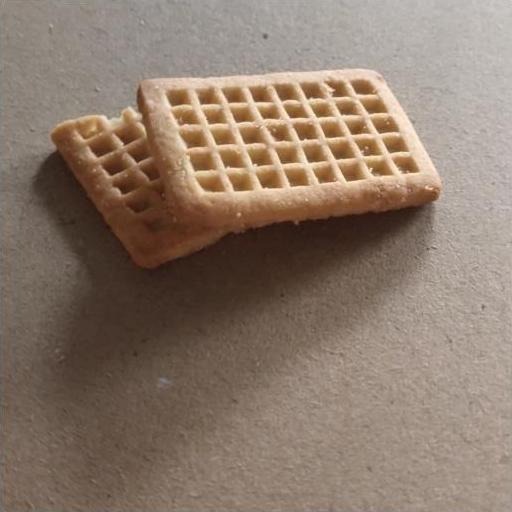} &
            \includegraphics[width=1\columnwidth]{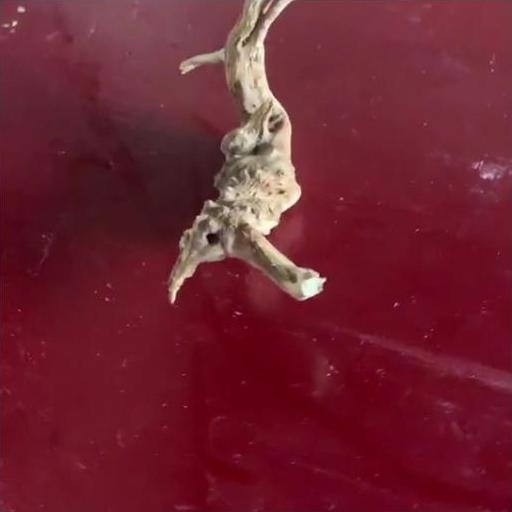} \\
        \end{tabular}%
    }

    \caption{MVImgNet failure cases. Each column represents one
    scene, while rows correspond to input view, predicted novel
    view, and target view.}
    \label{fig:mvimagenet_failure_cases_horizontal}
\end{figure}

\clearpage

\setcounter{section}{0}
\setcounter{figure}{0}
\setcounter{table}{0}
\setcounter{equation}{0}

\renewcommand{\thesection}{S\arabic{section}}
\renewcommand{\thefigure}{S\arabic{figure}}
\renewcommand{\thetable}{S\arabic{table}}
\renewcommand{\theequation}{S\arabic{equation}}

\renewcommand{\theHsection}{supp.\arabic{section}}
\renewcommand{\theHsubsection}{supp.\arabic{section}.\arabic{subsection}}
\renewcommand{\theHfigure}{supp.\arabic{figure}}
\renewcommand{\theHtable}{supp.\arabic{table}}

\twocolumn[
\section*{\centering{{Supplementary Material}}}
\vspace{10pt}
]
\section*{Additional Experimental Results}
We provide additional details and results on MVImgNet, including qualitative ablations and high-resolution evaluations. 

\section{MVImgNet}
\noindent \textbf{Pre-Processing:} We resize the shorter dimension of the images to be 512 and resize the other dimension to maintain the aspect ratio and then take centre crop of $512 \times 512$. These $512 \times 512$ RGB images are subsequently passed through VAE encoder and the DDIM inversion pipeline to get the inverted latents $\mathbf{z^\text{inv}}$ and extract their mean and variance components. We perform DDIM inversion from $t$ = 0 till $t$ = 600 in 30 steps. The data in the inverted latent space is of dimension $4\times64\times64$.

\noindent \textbf{Implementation Details:} TUnet has approximately 148M parameters. The dimensions of both class and camera embeddings are 64, and the cross-attention dimension is 768 with an attention head dimension of 64. We use the latent diffusion backbone \cite{Rombach_2022_CVPR}. For training, we randomly pair frames 1-10 of each scene with frames 15-25. We effectively use 20 frames per scene for training. We adopt the same frame pairing strategy for our evaluation. We train two models on our subset (i) 3 classes and (ii) 167 classes. Our 167 class model is trained for 450 epochs on a single 49 GB RTX A6000 for 14 GPU days with a batch size of 32 and a learning rate of 1e-5, and we decay the learning rate using a cycle scheduler.

\subsection{Synthesis using 167-class Model}
We show additional results using our 167-class model in Figure \ref{fig:grid_167_cls} . Our pipeline produces novel views that remain geometrically consistent with the targets while preserving global scene layout. We also see high texture fidelity and sharper high-frequency details.

\subsection{Synthesis of Multiple Views from Single Image }
\begin{figure*}[p]
    \centering

    \includegraphics[
        width=\textwidth,
        height=0.45\textheight,
        keepaspectratio
    ]{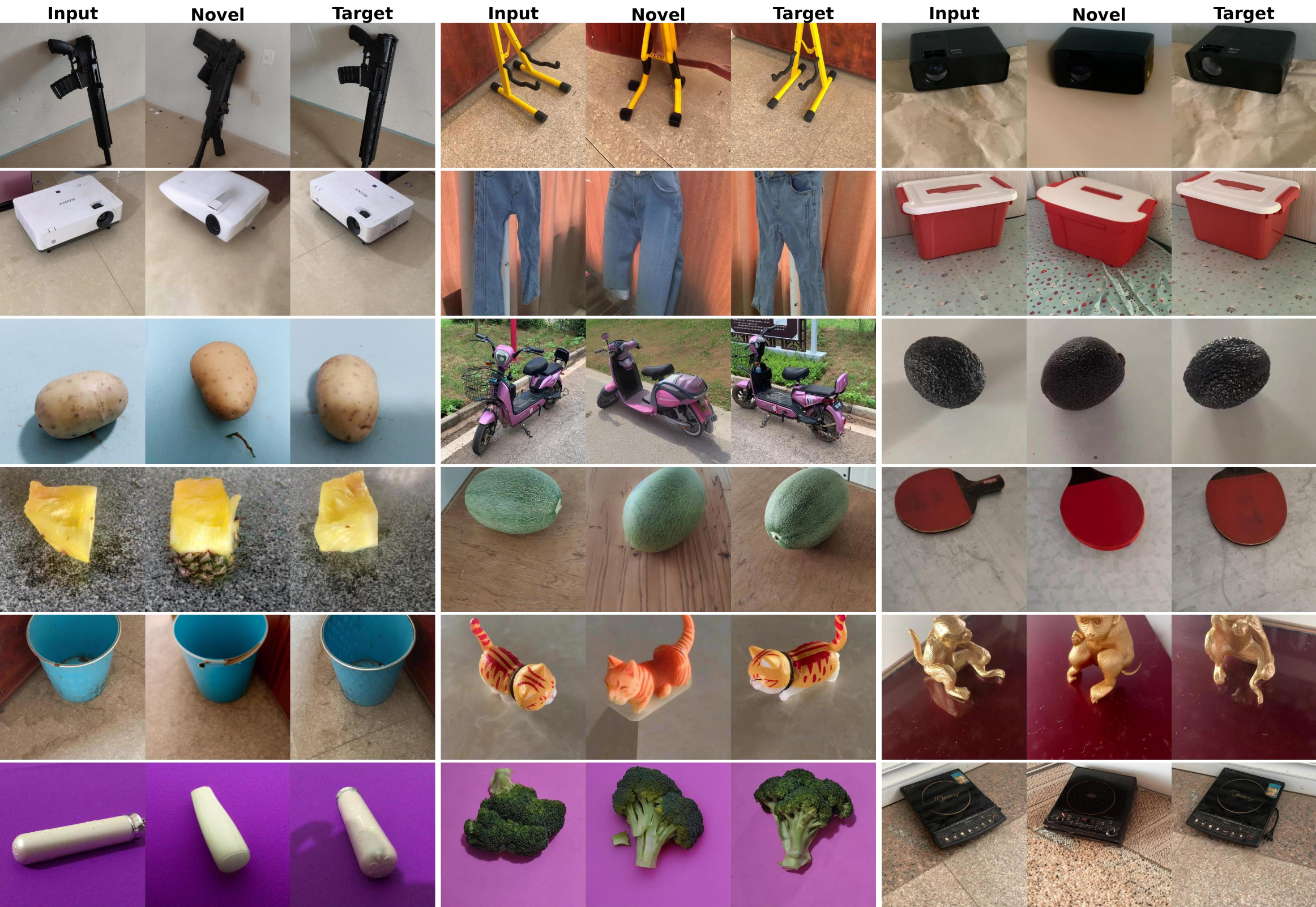}
    \caption{Additional Qualitative results with our 167-class trained model on MVImgNet test set.}
    \label{fig:grid_167_cls}

    \vspace{2em}

    \includegraphics[
        width=1\textwidth,
        height=0.48\textheight,
        keepaspectratio
    ]{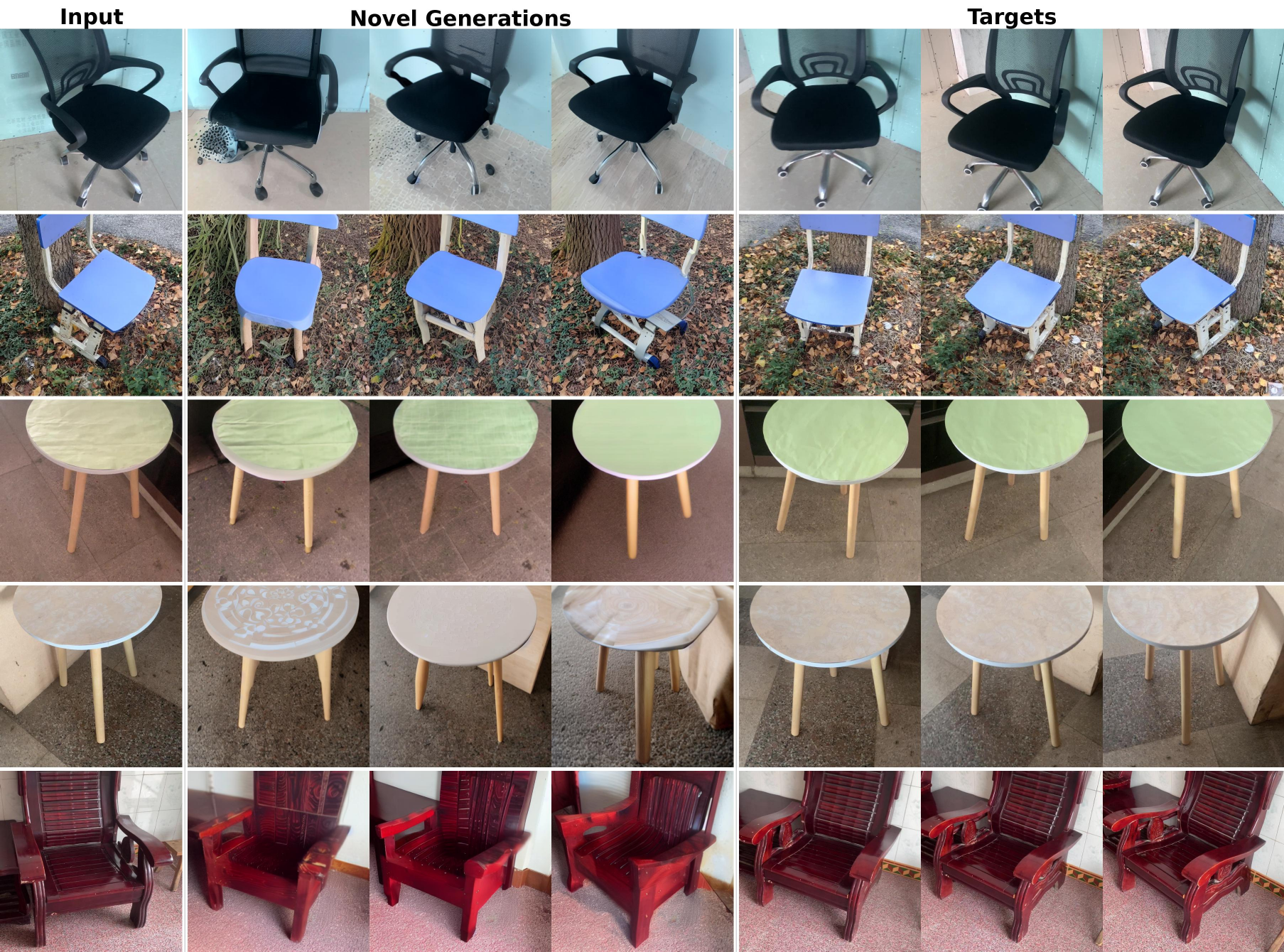}
    \caption{Generating multiple frames with single input image from MVImgNet.}
    \label{fig:multi_gen}
\end{figure*}

In Figure \ref{fig:multi_gen}, we generate multiple frames using a single input image. We query the same input image with multiple target camera parameters and reconstruct the novel views with respect to different target views. Even for long-range viewpoints, the proposed method achieves good synthesis results.

\subsection{Qualitative Ablation Results}
\begin{figure}[H]
    \centering
  \includegraphics[width=0.7\columnwidth,keepaspectratio]{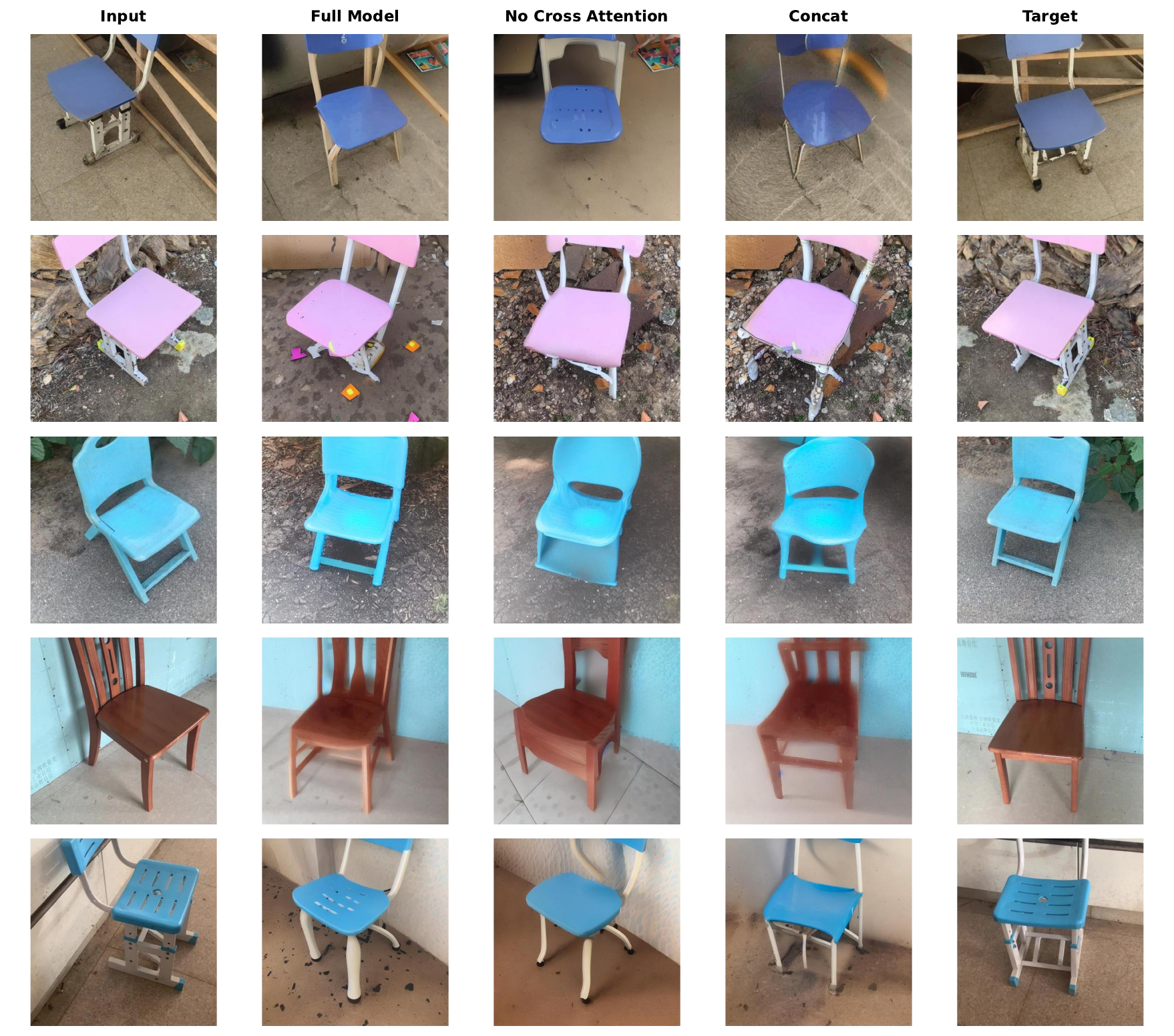}
    \caption{Qualitative ablation results. }
    \label{fig:ablation}
\end{figure}

In Figure \ref{fig:ablation}, we show qualitative results with our ablation setting. We can observe that in the `w/o Cross Attention' setting, the view transformation geometry suffers. In the `Concat' setting, the object loses fine-grained details while synthesizing the novel view. Full Model shows the best performance.

\subsection{Additional Results}

In Table \ref{tab:our_512_results}, we report the results on the original synthesized resolution of $512\times512$.

\noindent \textbf{3D Consistency Evaluation (MEt3R).}
To evaluate geometric consistency between the input view and synthesized novel views, we compute the MEt3R metric for each input-target pair across the test set of 18,358  source–target pairs using the official implementation of MEt3R~\cite{asim25met3r}. 
Lower values indicate stronger multi-view geometric consistency. 
As shown in Table~\ref{tab:met3r_results}, the MEt3R score gradually increases as the frame distance between input and target views grows. 
This trend is expected since larger viewpoint changes introduce stronger occlusions and require the model to synthesize previously unseen surfaces. 
Nevertheless, the increase remains moderate across the full range (from $0.432$ to $0.495$), suggesting that the proposed framework maintains stable geometric structure even under long-range viewpoint transformations.

\begin{table}[!htbp]
    \centering
    \small
    \setlength{\tabcolsep}{4pt}
    \renewcommand{\arraystretch}{1.0}
    \captionsetup{position=top,skip=3pt}

    \caption{Our results at original resolution of $512 \times 512$}
    \label{tab:our_512_results}

    \begin{tabular}{@{}lcccc@{}}
        \toprule
        Method & LPIPS $\downarrow$ & PSNR $\uparrow$
               & SSIM $\uparrow$ & FID $\downarrow$ \\
        \midrule
        Ours & 0.556 & 16.162 & 0.578 & 69.604 \\
        \bottomrule
    \end{tabular}

    \par\vspace{6pt}

    \caption{MEt3R geometric consistency between the input view
    and synthesized novel views across different frame distances.
    Lower values indicate stronger multi-view geometric alignment.}
    \label{tab:met3r_results}

    \begin{tabular}{@{}lccc@{}}
        \toprule
        Frame Distance & $N$ & MEt3R (mean)$\downarrow$ & Std. \\
        \midrule
        $\leq 5$ & 196  & 0.4319 & 0.1827 \\
        6--12    & 5875 & 0.4621 & 0.2283 \\
        13--18   & 8801 & 0.4787 & 0.2650 \\
        19--25   & 3486 & 0.4945 & 0.2853 \\
        \bottomrule
    \end{tabular}
\end{table}

\clearpage

{\small
\bibliographystyle{ieeenat_fullname}
\bibliography{main}
}

\end{document}